\documentclass[10pt]{article}

\usepackage[letterpaper,left=0.5in,right=0.5in,top=0.75in,bottom=0.80in]{geometry}
\usepackage[hidelinks]{hyperref}
\usepackage{natbib}
\setcitestyle{numbers,sort&compress,open={[},close={]},citesep={,}}
\usepackage{balance}
\title{VEPHand: View-Efficient Photometric Hand Performance Capture at Scale}

\author{
    \begin{tabular}{c}
        Zhengyang Shen \quad Kai-Hung Chang \quad Erroll Wood \quad Deying Kong \quad Bo Peng \\
        Timo Bolkart \quad Jinlong Yang \quad Bowen Zhao \quad Danhang Tang \quad Sasa Petrovic \quad Emre Aksan \\
        Jérémy Riviere \quad Vassilis Choutas \quad Delio Vicini \quad Jay Busch \quad Shichen Liu \\
        Zhe Cao \quad Hugh Liu \quad JingJing Shen \quad Jonathan Taylor \quad Mingsong Dou \\[2em]
        \Large{Google XR}
    \end{tabular}
}
\date{}
\usepackage{url}
\usepackage{multirow}
\usepackage{xcolor,colortbl}
\usepackage{graphicx}

\usepackage{dblfloatfix}

\usepackage{microtype}

\usepackage{amsmath}
\usepackage{amssymb}
\usepackage{booktabs}

\usepackage{enumitem}
\usepackage{graphicx} 
\usepackage{arydshln} 
\usepackage{svg}
\usepackage{subcaption} 
\usepackage{capt-of} 
\usepackage[autostyle, english = american]{csquotes}
\MakeOuterQuote{"}

\usepackage{titlesec}
\titleformat{\section}
  {\normalfont\large\bfseries}{\thesection}{0.75em}{}
\titleformat{\subsection}
  {\normalfont\normalsize\bfseries}{\thesubsection}{0.75em}{}

\begin{document}

\twocolumn[
\vspace{-3.0em}
  \maketitle
  \vspace{-2.0em}
  \begin{abstract}
  Robust, high-fidelity 3D hand capture, while fundamental to digital human creation, remains challenging with practical multi-view systems that balance rich photometry with the geometric ambiguities of reconstruction arising from limited viewpoint density. This paper presents an end-to-end pipeline for dynamic hand performance capture and registration, specifically designed for view-efficient setups ($\sim$20 views). We address key challenges with two primary innovations. First, to overcome reconstruction difficulties like limited view overlap and background clutter, our mask-free neural method robustly extracts detailed hand geometry and appearance from unmasked images using scene parameterization and scenario-specific density regularization. Second, addressing registration challenges such as accurately capturing non-linear skin deformations and ensuring plausible results during severe self-contact, we propose a physics-inspired framework. It aligns reconstructions to a personalized hand model by optimizing intrinsic volumetric offsets within its canonical tetrahedral mesh, alongside pose parameters. This approach, supported by robust losses and optimization, captures fine surface deformations, ensures plausible results under severe articulation and self-contact, and demonstrates strong tolerance to input noise. We demonstrate the scalability and robustness of our automated pipeline on an extensive dataset of over 12,000 sequences, from which we also derive a large-scale, high-quality synthetic 2D/3D hand dataset for training downstream tasks. This showcases its effectiveness for single hands, intricate two-hand interactions, and natural hand-object manipulations. Our method achieves state-of-the-art reconstruction fidelity in view-efficient, unmasked scenarios and highly accurate registration (approx. 1mm mean surface error), significantly advancing practical, high-quality hand capture.
Our project page are available at \url{https://vephand.github.io/}.
  \end{abstract}


  \vspace{0.75em}
  \refstepcounter{figure}\label{fig:teaser}
  {\centering
    \includegraphics[width=\textwidth]{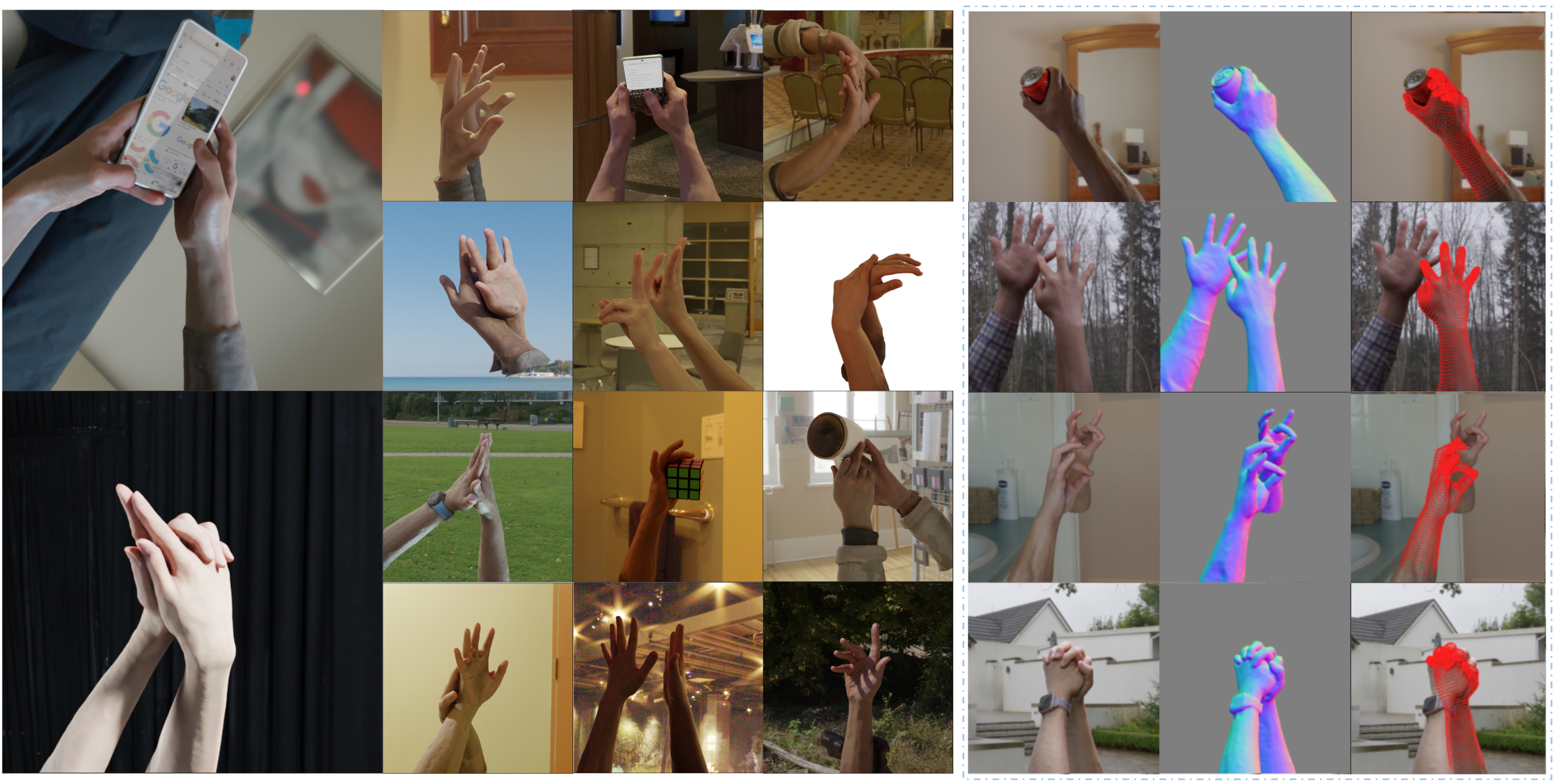}\par
    {\small\textbf{Figure \thefigure.} VEPHand delivers high-fidelity, versatile digital hands at scale. The left panel showcases photorealistic renderings from our large-scale dataset, demonstrating robustness across diverse scenarios including complex single-hand poses, intimate two-hand interactions, and natural hand-object manipulations (e.g., using a phone, grasping a can). The right panel (blue dashed box) provides a technical visualization of our pipeline's accuracy: for each example, we show the rendered appearance, the detailed surface normals recovered by our neural reconstruction, and the precise alignment of our volumetric registration model (red wireframe overlay) to the captured performance.\par}
  }

  \vspace{1.0em}
]




\section{Introduction}

The human hand, with its intricate structure and astonishing dexterity, is fundamental to how we interact with the world. Capturing its nuanced motion in high fidelity is a cornerstone of modern computer graphics, vital for creating expressive digital humans, advancing robotics, and enabling immersive virtual and augmented reality experiences. While specialized studios with dense camera arrays~\cite{joo2015panoptic}, active depth sensors ~\cite{qian2014realtime, supanvcivc2018depth} or intrusive markers~\cite{endo2014reconstructing} can produce exceptional results, their cost and complexity make them impractical for large-scale data acquisition. To address this, we introduce a unified hardware-software framework built upon a purely vision-based perception setup. Our approach is designed to navigate the trade-offs between camera density, capture diversity, and reconstruction precision, enabling high-fidelity hand modeling from efficient, practical multi-view configurations.

The pursuit of view efficiency exposes the limitations of established 3D reconstruction paradigms. Traditional Multi-View Stereo (MVS) struggles to find reliable feature correspondences across the large baselines and severe occlusions inherent in relative sparse-view hand capture. More recent neural techniques~\cite{mildenhall2021nerf}, while powerful, also face critical hurdles. Methods based on Signed Distance Functions (SDFs)~\cite{yariv2020multiview, wang2021neus, oechsle2021unisurf, yariv2023bakedsdf}, a mainstream choice for high-quality surfaces, often depend on accurate foreground masks, a dependency that proves brittle and unreliable for complex, dynamic hands at scale. While mask-free SDF methods exist, our extensive experiments (Sec.~\ref{sec:study_recons}) show they frequently fail to converge or produce incomplete geometry in our unconstrained setting. Similarly, emerging techniques like 3D Gaussian Splatting~\cite{kerbl20233d} (GS) and its 2D variant~\cite{huang20242d} for surface extraction, while powerful for rendering, require robust initialization and extracting high-quality meshes remains a non-trivial challenge. As a result, enabling high-fidelity capture in these practical settings requires a reconstruction approach specifically tailored to navigate the geometric ambiguities inherent to sparse, unmasked photometric systems.

To overcome these reconstruction challenges, we introduce our first primary innovation: a novel, mask-free neural reconstruction method that robustly extracts detailed hand geometry and appearance from unmasked, view-efficient imagery. Instead of relying on an SDF, we employ direct density prediction, a choice that proved critical for achieving stable convergence across our large-scale dataset of diverse subjects and interactions. We tackle the ambiguity of unmasked inputs by integrating scene parameterization with a crucial, scenario-specific density regularization framework. This tailored regularization actively suppresses common artifacts like "floaters" and prevents background collapse, enabling the network to form sharp, topologically sound surfaces with a remarkably low failure rate. Our tailored mesh extraction and texturing process (Sec. \ref{sec:meshing}) allows us to recover high-quality surface appearance directly from the learned density field. This approach provides the robust, high-quality geometric data that is the foundation for our scalable pipeline.

Robust reconstruction alone is not enough; for animation and analysis, this geometry must be aligned to a consistent parametric model. Our second key innovation is a physics-inspired volumetric registration framework designed to capture these fine details automatically. We first generate a unique Personalized Hand Template (PHT) for each subject, which includes a consistent underlying tetrahedral mesh. A key insight of our approach is to separate the deformation into two complementary components. We leverage standard, pre-trained pose-dependent correctives (similar to those in MANO~\cite{romero2022embodied}) to handle broad, Linear Blend Skinning (LBS)-related artifacts. The fine performance-specific details are then captured by optimizing a set of {\it intrinsic volumetric offsets} applied directly to the PHT's canonical tetrahedral mesh. This canonical-space deformation, guided by volumetric regularizers, ensures that these detailed adjustments are volume preserved. This dual-component strategy allows us to benefit from the power of existing parametric models while robustly fitting the intricate surface dynamics present in our reconstructions, achieving highly accurate alignment (approx. 1mm surface error).

In this paper, we present VEPHand, a comprehensive, end-to-end pipeline that integrates these innovations into a scalable system for View-Efficient Photometric Hand performance capture. We demonstrate its effectiveness by processing an extensive dataset of nearly 12,000 sequences from 800 subjects, showcasing its robustness across single-hand, two-hand, and hand-object interactions. Our contributions are:

 \textit{A View-Efficient Photometric Capture System ("Handbooth"):} We built a 20-camera photometric capture system engineered for view-efficient, high-quality hand performance data acquisition. It features strategic camera placement on a custom rig, a heterogeneous lens strategy and synchronized high-resolution recording, providing the rich photometric input crucial for our reconstruction and registration pipeline.
 
 \textit{Scalable Non-SDF Neural Reconstruction for View-Efficient Setups:} A novel neural rendering technique that achieves high geometric fidelity in view-efficient settings, demonstrating superior convergence and robustness over common SDF-based alternatives.
 
 \textit{High-Fidelity Volumetric Physics-Inspired Registration:} A novel registration method captures fine-grained, non-linear surface details by optimizing intrinsic canonical offsets on a personalized tetrahedral mesh under physics-inspired constraints.
 
 \textit{Scalable, Automated, and Validated End-to-End Pipeline:} A fully automated pipeline for high-fidelity hand capture at this scale, validated by processing a massive dataset with a single, fixed set of parameters, and the creation of a large-scale synthetic dataset and an improved parametric hand model from its outputs.
 

The paper is organized as follows: We first review related work (Sec.~\ref{sec:related_work}) and describe our Handbooth capture system (Sec.~\ref{sec:handbooth_system}). We then detail our neural reconstruction (Sec.~\ref{sec:reconstruction}) and volumetric registration (Sec.~\ref{sec:registration}) methods. Subsequently, we present extensive results and evaluations (Sec.~\ref{sec:results}), discuss applications (Sec.~\ref{sec:application}), and conclude with a discussion of limitations and future directions (Sec.~\ref{sec:discussion}).

\section{Related Work}
\label{sec:related_work}

Capturing and modeling the intricate dynamics of human hands with high fidelity, especially at scale, remains a central challenge in computer graphics and vision, pivotal for applications from immersive digital humans to robotics. Our work, VEPHand, builds upon and extends a rich history of research across several key domains. This section reviews prior art in hand performance capture systems, neural scene reconstruction techniques (with a particular focus on dynamic hands and the challenges of scalability), methods for parametric hand modeling and registration, and approaches to scalable data processing alongside the evolution of hand datasets. By examining existing methodologies and their inherent limitations, particularly concerning automation and large-scale deployment with practical capture setups, we contextualize the contributions of VEPHand towards achieving robust, scalable, and high-fidelity hand performance capture.

\subsection{Hand Performance Capture Systems}
\label{sec:rw_capture_systems}

A variety of systems have been developed for capturing hand performance, each with distinct trade-offs regarding accuracy, intrusiveness, cost, and scalability.

\paragraph{Marker-based Motion Capture.} These systems, exemplified by Vicon or OptiTrack setups used in datasets like ARCTIC~\cite{fan2023arctic} and GRAB~\cite{taheri2020grab}, attach physical markers to the hand to track 3D joint positions with high accuracy and temporal resolution. While considered a gold standard for precise kinematics, they are intrusive, can alter natural motion, suffer from marker occlusion (especially for hand self-occlusion), require complex setups, and necessitate laborious post-processing. Their primary modern role is often in generating ground truth for training and evaluating markerless methods.

\paragraph{Systems with Active Sensors.} RGB-D cameras (e.g., Kinect, RealSense, Azure Kinect) and other active sensors (e.g., Leap Motion) directly measure depth, simplifying 3D reconstruction~\cite{mueller2017real, sridhar2015fast}. Datasets like HOGraspNet~\cite{cho2024dense} (Azure Kinect) and HOnnotate~\cite{hampali2020honnotate} (Kinect) leverage this. Recently, PALM~\cite{fan2025palm} utilized a multi-view active stereo photogrammetry system with light projectors to acquire high-fidelity single-hand geometry. While active systems offer robustness to textureless surfaces, they face limitations regarding range, material sensitivity (e.g., dark or shiny surfaces), and multi-sensor interference. In contrast, our pure photometric system circumvents these limitations through a view-efficient design, utilizing 20 cameras placed uniformly to maximize photometric efficiency. This enables the robust capture of much more diverse two-hand interactions, even under strong occlusion.

\paragraph{Monocular Systems.}
Utilizing a single RGB camera~\cite{zimmermann2019freihand, boukhayma20193d,li2022interacting, huang2022reconstructing} or depth sensor offers maximum accessibility and low cost, driving significant research. Early learning-based methods directly regressed 3D joint locations~\cite{zimmermann2017learning}, while subsequent works advanced to jointly estimating 3D hand shape and pose, often by predicting parameters for models like MANO~\cite{romero2022embodied, baek2019pushing, yang2020bihand, wang2020rgb2hands} or, more recently, leveraging transformer architectures~\cite{lin2021end, pavlakos2024reconstructing}. Capturing interacting hands from monocular RGB presents even greater occlusion challenges, with datasets like InterHand2.6M~\cite{moon2020interhand2} (originally multi-view) spurring progress~\cite{meng20223d}. Some approaches, like HARP~\cite{karunratanakul2023harp}, use short monocular video sequences to reconstruct personalized appearance. However, inferring 3D from a single viewpoint is inherently ill-posed. Despite advancements, fundamental limitations due to occlusion and depth ambiguity make it challenging for purely monocular methods to consistently recover accurate, fine-scale geometric details, especially for complex interactions, thus motivating the need for multi-view systems for higher fidelity capture.

\paragraph{Multi-view Photometric Systems.} These systems utilize synchronized camera arrays to capture markerless hand motion, leveraging multi-view stereo (MVS) and photometric consistency. They are generally categorized by the density of the capture rig:
\begin{enumerate}[label=\arabic*), itemsep=0pt, topsep=1pt, parsep=1pt]
\item \textbf{Dense and Active Systems.} High-end configurations, such as the Panoptic Studio~\cite{joo2018total} (100+ cameras) or commercial active stereo scanners (e.g., 3dMD used in~\cite{fan2025palm}), define the gold standard for reconstruction fidelity. While they offer exceptional detail, they face distinct limitations. Dense passive arrays impose extreme costs and data management hurdles, restricting them to specialized laboratories. Meanwhile, active stereo solutions, despite their precision, often struggle with complex interactions; severe occlusions can block the projected structured light patterns required for depth estimation, limiting their effectiveness to open or pre-defined poses.
\item \textbf{Sparse/Moderate-view Systems.} To improve practicality, systems with fewer cameras (typically 4--20, e.g., FreiHAND~\cite{zimmermann2019freihand} or~\cite{pang2024sparse, zheng2023hamuco, zhao2020hand,chen2021mvhm, rim2026show3d}) are increasingly favored. While these setups offer stronger geometric constraints than monocular methods, they expose the fragility of traditional reconstruction algorithms. The wide baselines and frequent self-occlusions inherent to sparse hand capture cause standard MVS to fail, resulting in noisy or incomplete geometry. Consequently, recovering high-fidelity surface details robustly from such view-efficient setups, without relying on brittle masking, remains a critical open challenge.
\end{enumerate}

Our \textit{Handbooth} system (20 cameras) targets this moderate regime, enabling the capture of diverse, unconstrained hand poses with a significantly reduced hardware footprint. While this view-efficient design improves scalability, it imposes severe algorithmic challenges due to limited overlap. VEPHand addresses these specifically, employing tailored regularization and registration to extract high-fidelity geometry where standard methods struggle.

\subsection{Neural Scene Reconstruction for Dynamic Hands}
\label{sec:related_work_neural_reconstruction} 

The advent of Neural Radiance Fields (NeRF) \cite{mildenhall2021nerf} revolutionized 3D scene representation from images. However, adapting NeRF for high-fidelity, dynamic hand capture from view-efficient, unmasked inputs presents significant challenges that are not fully addressed by foundational improvements like mip-NeRF \cite{barron2021mip, barron2022mip} and Instant NGP \cite{muller2022instant} or dynamic extensions \cite{pumarola2021d, park2021nerfies, park2021hypernerf, fridovich2023k}.

\paragraph{Surface-focused Representations and the Limitations of SDFs.} To achieve precise surface geometry, many leading neural rendering techniques incorporate Signed Distance Functions (SDFs). These methods leverage strong geometric priors to produce smooth, detailed surfaces. However, for scalable and unconstrained hand capture, SDF-based approaches face critical limitations in our target domain. First, many require or benefit significantly from accurate foreground masks~\cite{wang2021neus,yariv2021volume}. Automating mask generation at scale for dynamic hands is notoriously difficult; our experiments with state-of-the-art segmentation models revealed inconsistencies that could degrade, rather than improve, reconstruction quality. Second, and more fundamentally, SDF optimization can be sensitive in sparse-view, unmasked settings. Without strong multi-view consensus or mask supervision, these methods can struggle to converge, lose fine details, or produce incomplete surfaces, as the inherent smoothness priors (e.g., Eikonal loss) can be detrimental when capturing sharp details or thin structures~\cite{patel2024normal}. Our evaluation against prominent SDF-based methods like BakedSDF~\cite{yariv2023bakedsdf}, UniSDF~\cite{wang2024unisdf}, and Neuralangelo~\cite{li2023neuralangelo} confirms this, showing high failure rates on our challenging dataset (Sec 6.2). These findings motivated our choice to pursue an alternative representation. VEPHand is built upon direct density prediction, which, combined with our tailored regularization, proved significantly more robust and achieved superior convergence for this task.

\paragraph{Emerging Representations like 3D Gaussian Splatting.} While 3D Gaussian Splatting (3DGS)~\cite{kerbl20233d} is powerful, it depends on robust SfM initialization, which is unreliable in our limited overlap setup. Although some variants~\cite{huang20242d, guedon2025matcha, fan2024instantsplat} offer better stability and performance, we find they frequently result in over-smoothed geometry lacking high-frequency detail. Furthermore, specialized methods~\cite{pokhariya2024manus, dong2025handsplat, xiao2025rogsplat, qu2025hogsa} rely on pre-computed parametric model tracking (e.g., MANO~\cite{romero2022embodied} and SMPL-X~\cite{pavlakos2019expressive}). This introduces a circular dependency: our primary objective is to generate the high-fidelity ground truth required to train such trackers. To break this cycle, we design our system to operate \textit{ab initio}, reconstructing detailed geometry and appearance directly from unmasked imagery without external parametric priors. By establishing this robust, ground-truth-independent foundation, we provide a scalable framework that can subsequently leverage these advanced tracking and representation methods to further augment performance.

These challenges with existing paradigms motivate VEPHand's non-SDF reconstruction pipeline. While early non-SDF works like Neural Volumes~\cite{lombardi2019neural} or Plenoxels~\cite{fridovichkeil2022plenoxels} demonstrated potential, achieving consistent, high-fidelity surfaces without an SDF's explicit guidance remained an open problem. Our work contributes a specialized solution for hands, using a combination of direct density prediction and innovations like scenario-specific regularization to achieve SDF-level surface quality with the enhanced robustness required for scalable capture from sparse, unmasked imagery.

\subsection{ Scalable Hand Data Processing and Existing Hand Datasets}
The advancement of robust 3D hand pose estimation and interaction understanding is critically dependent on large-scale, diverse, and accurately annotated datasets. Early datasets (e.g., STB \cite{zhang20163d}, NYU Hand Pose \cite{tompson2014real}) were often limited in scale or diversity due to laborious manual annotation or reliance on specific sensor modalities. Recognizing these limitations, the field has seen a significant push towards more scalable data acquisition and annotation methodologies.

A key trend has been the use of multi-view capture systems combined with semi-automated or fully-automated annotation pipelines. Datasets like FreiHAND \cite{zimmermann2019freihand} demonstrated scalable annotation via human-in-the-loop refinement of model-based fits from multi-view green-screen captures. InterHand2.6M \cite{moon2020interhand2} achieved massive scale by leveraging a very dense multi-view studio to enable highly accurate machine annotation for interacting hands. Other works have utilized marker-based motion capture for high-accuracy ground truth, particularly for complex bimanual and articulated object interactions (e.g., ARCTIC \cite{fan2023arctic}), followed by fitting parametric models. Datasets focusing on rich interactions, such as HOGraspNet~\cite{cho2024dense} (grasp taxonomy) or HOI4D \cite{liu2022hoi4d} (category-level egocentric interaction), also employ combinations of multi-view capture, sensor fusion, and sophisticated semi-automated annotation. More recently, efforts like GigaHands \cite{fu2024gigahands} and TACO \cite{liu2024taco} showcase fully automated pipelines achieving unprecedented scale by integrating advanced AI-driven perception and reconstruction.

However, a common thread among many of these seminal datasets is a reliance on either dense camera setups, intrusive markers, or a primary focus on skeletal pose rather than detailed, textured surface geometry. A significant gap remains in the scalable acquisition of high-fidelity surface data from practical, view-efficient photometric systems, which presents unique challenges in robust reconstruction and registration. 

\subsection{Hand Registration and Parametric Modeling}
\label{sec:related_work_registration} 

Accurate registration of captured hand performances to a common representation is essential for animation, analysis, and creating personalized digital assets. This process typically involves aligning observations with hand models and employing sophisticated registration techniques to handle the hand's complexity.

\paragraph{Parametric Hand Models.} Statistical models like MANO~\cite{romero2022embodied} and the hand component of SMPL-X~\cite{pavlakos2019expressive} have become influential, offering low-dimensional representations of hand shape and pose learned from 3D scan datasets. Further efforts to enhance parametric hand model realism include incorporating more anatomical details, such as in NIMBLE \cite{li2022nimble}, which explicitly includes volumetric muscle groups alongside bones and skin to achieve more anatomically plausible animations. While these models primarily encode detailed deformations into learned blendshapes derived from offline registration , advancements like Handy extend this generalized framework by incorporating a large-scale dataset of over 1,200 subjects to capture universal shape variations across a broad demographic. Similarly, GHUM~\cite{xu2020ghum} provide a generalized, holistic modeling approach, utilizing non-linear variational autoencoders to capture complex correlations between full-body and hand articulation. In contrast, some methods prioritize high-fidelity subject-specific detail; DeepHandMesh~\cite{moon2020deephandmesh} is designed for a personalized environment, training its encoder-decoder framework on a single subject to replicate realistic details such as creases and skin bulging. Bridging these paradigms, PHRIT~\cite{huang2023phrit} utilizes a novel deformation field to combine parametric meshes with implicit templates, enabling efficient, high-fidelity reconstruction at infinite resolution. The Universal Hand Model (UHM)~\cite{moon2024authentic} bridges these concepts by providing a generalized model capable of representing arbitrary identities that can be efficiently adapted into a personalized, authentic avatar using only a short monocular phone scan. The development of such models highlights the importance of learning from large, diverse registered datasets to capture population variance and complex deformations~\cite{qian2020html}. 
Our VEPHand registration pipeline builds upon these advancements, starting from and later contributing to our own high-resolution parametric hand template (Sec.~\ref{sec:parametric_model})

\paragraph{Registration Techniques.} Aligning hand models to observations has evolved significantly.
Classical geometric alignment methods like Iterative Closest Point (ICP)~\cite{besl1992method, rusinkiewicz2001efficient} are effective for rigid objects but struggle with the partial data, non-rigid deformations, and self-occlusions common in hand capture.
Optimization-based fitting is a dominant paradigm, where model parameters are optimized to match various cues, including 2D keypoints~\cite{zimmermann2017learning}, silhouettes~\cite{sridhar2014real}, depth maps~\cite{tkach2016sphere}, or 3D scans~\cite{bogo2014faust}. These methods often integrate learned priors or deep features to improve robustness.

Physics-inspired and volumetric regularization have become crucial for generating plausible deformations and handling complex phenomena like self-contact. Techniques like As-Rigid-As-Possible (ARAP) surface modeling~\cite{sorkine2007rigid} preserve local shape details. More advanced approaches employ Finite Element Methods (FEM) with hyperelastic material models (e.g., Neo-Hookean~\cite{smith2018stable}) to simulate soft tissue behavior and volumetric integrity.

A notable example~\cite{smith2020constraining} achieves robust tracking by directly optimizing the world-space vertex positions of a posed volumetric tetrahedral mesh, regularized by an elastic energy term relative to a single rest state. While powerful, this approach faces scalability challenges in per-subject mesh generation and applies physical constraints to the full, posed mesh. VEPHand offers a distinct formulation designed for both detail and scalability. We propose a dual-component deformation model that separates concerns: {\it Low-frequency, pose-driven deformations} are handled by standard corrective blendshapes, leveraging the power of established parametric models. {\it Fine-grained, performance-specific details} are captured by optimizing a set of intrinsic volumetric offsets on the \textit{canonical} tetrahedral mesh of a personalized template, ensuring physical plausibility before global posing.

\begin{figure}
    \centering
    \includegraphics[width=\linewidth]{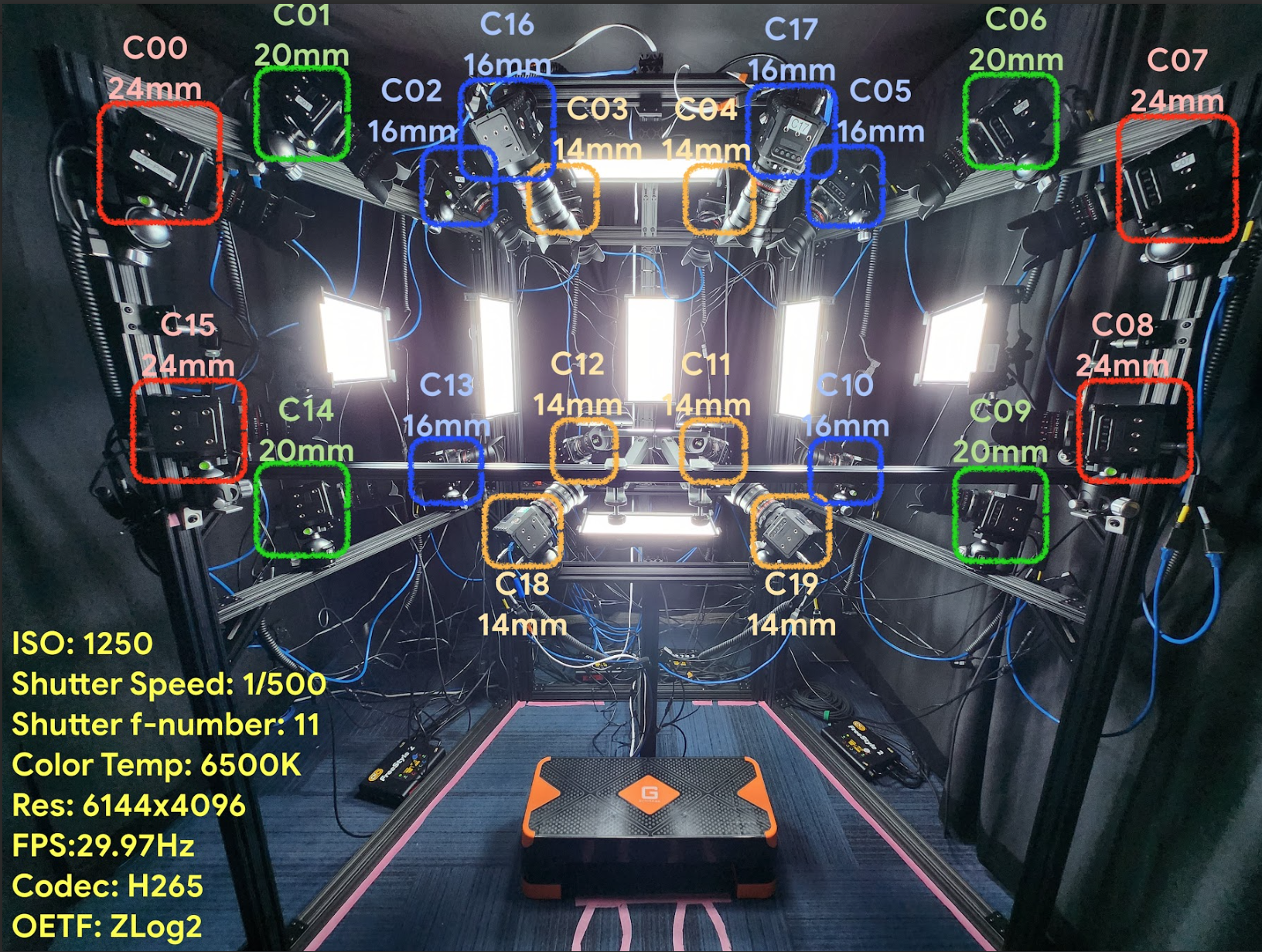}
    \caption{ Our Handbooth system configured for general hand performance capture. The 20 cameras utilize a heterogeneous lens strategy (focal lengths indicated: 24mm [red], 16mm [blue], 20mm [green], 14mm [orange]) to balance field of view and resolution on the hand surface from different positions within the custom triple-panel rig. Diffuse illumination is provided by large panel lights. The narrow Field-of-View (FoV) configuration can be found in Supplementary Material. (Fig.~\ref{fig:handbooth_small_fov_setup}).
}
    \label{fig:handbooth_wide_fov_setup}
\end{figure}

\begin{figure*}
    \centering
    \includegraphics[width=\linewidth]{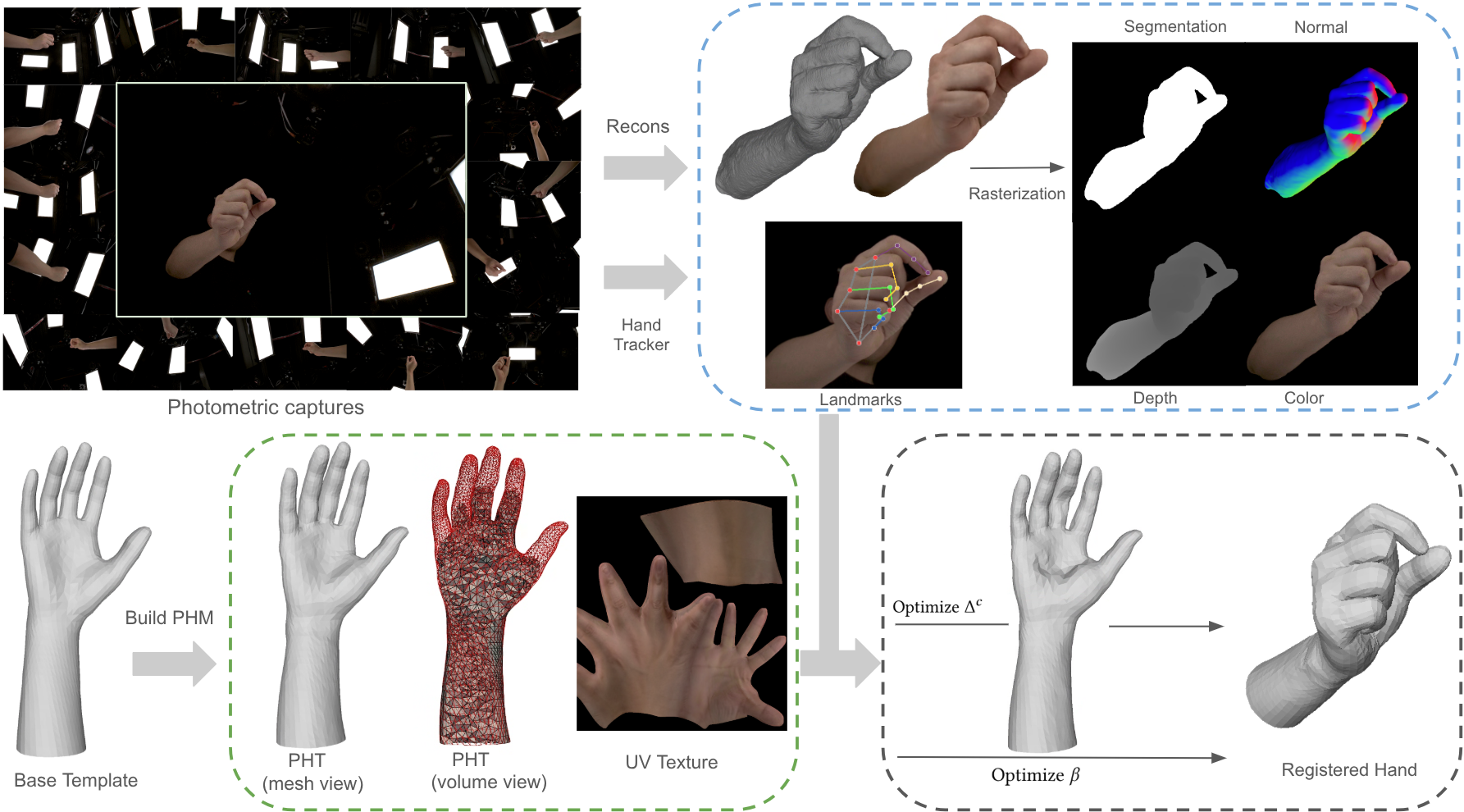}
    \caption{The Handbooth data processing pipeline. Multi-view photometric captures (top-left) from our system (Sec.~\ref{sec:handbooth_system}) are initially processed to generate 3D reconstructions ("Recons", Sec.~\ref{sec:reconstruction}) and estimate 3D hand landmarks ("Hand Tracker", Appendix \ref{sec:landmark_estimation}). Subsequently, a generic template hand is adapted to each subject to create a  Personalized Hand Template (PHT) (Sec.~\ref{sec:PHT_generation}, green box), which features a surface mesh, an underlying volumetric tetrahedral mesh (sliced view) as described in our deformation model (Sec.~\ref{sec:deform_model}), and a UV texture(Sec.~\ref{sec:PHT_generation}). For each frame, this PHT is registered to the captured data (Sec.~\ref{sec:registration}, black box) by optimizing pose parameters ($\beta$) and intrinsic canonical volumetric offsets ($\Delta^c$). This registration process is guided by the 3D reconstructions, landmarks, and various 2D data modalities (segmentation, normal, depth, and color maps obtained via rasterization for comparison, blue box), which are incorporated through our data fidelity terms (Sec.~\ref{sec:data_term}), ultimately yielding the final registered hand.}
    \label{fig:handbooth}
\end{figure*}

\section{Handbooth: A System for View-Efficient Photometric Hand Capture}
\label{sec:handbooth_system}
Our hand performance data is acquired using "Handbooth," a custom-designed multi-view capture system, shown in Fig.~\ref{fig:handbooth_wide_fov_setup}. A central characteristic of this system is its reliance exclusively on photometric information captured by a set of synchronized 20 Z-Cam cameras (model E2-S6Gs). The cameras are mounted on a custom triple-panel, semi-open rig constructed from aluminum extrusions, providing hemispherical viewpoint coverage while allowing subject access.

Strategically, cameras can be equipped with prime lenses selected to suit the capture objective. For general motion capture prioritizing larger hand movements (our primary configuration), we use lenses of varying focal lengths (e.g., 14 - 24mm), assigning longer focal lengths to farther cameras and wider lenses to closer ones. This heterogeneous optical setup aims to maintain relatively consistent pixel density on the hand surface across all views despite the limited number of viewpoints.

For scenarios focused on capturing fine-scale surface appearance and micro-details, we adapt the system to a narrow Field-of-View (FoV) configuration, as illustrated in Fig.~\ref{fig:handbooth_small_fov_setup}. This typically involves equipping most, if not all, cameras with narrower focal length lenses (e.g., a combination of 14mm, 16mm, 20mm, and 24mm lenses). This setup magnifies the hand surface for texture detail but inherently reduces view overlap and the capture volume. These characteristics present specific geometric challenges that necessitate tailored processing, such as our $L_{\text{contain}}$ regularization (Sec.~\ref{sec:density_regularization}).

Controlled illumination is provided by multiple large box lights positioned to create soft, diffuse lighting within the capture volume (approx. 150cm x 100cm x 70cm), minimizing harsh shadows. Precise temporal synchronization across all cameras is achieved using a dedicated hardware triggering signal initiated by a “master” camera and propagated sequentially to all “client / slave” cameras via specialized synchronization cable (Z Cam E2). Systematic calibration yields accurate intrinsic parameters for each camera-lens pair and the joint extrinsic parameters for all cameras in a common coordinate system. This is achieved by capturing a calibration pattern from multiple viewpoints with several cameras simultaneously, followed by a bundle adjustment algorithm. This process yields high accuracy, with a resulting mean reprojection error of less than 1 pixel. The system captures synchronized video at 29.97 Hz and 6144x4096 pixels using the ZLog2 profile for enhanced dynamic range. The raw camera data undergoes decoding, linear color space conversion, and undistortion with calibrated parameters. The final input for our pipeline is a sequence where each frame consists of 20 synchronized, undistorted images of different views, as well as the corresponding camera calibration for a given capture session.

\section{Reconstruction}
\label{sec:reconstruction}
\noindent Our approach integrates four key components to achieve high-quality results: a robust mask-free foundation combining scene parameterization with surface refinement (Sec.~\ref{sec:reconstruction_model}), a scenario-specific density regularization to enforce geometric consistency (Sec.~\ref{sec:density_regularization}), accelerated training strategies (Sec.~\ref{sec:acc_train}), and a tailored mesh extraction and texturing process (Sec.~\ref{sec:meshing}).

\subsection{Mask-Free Density Field and Surface Refinement}
\label{sec:reconstruction_model}
To avoid the fragility of mask dependency, we adopt a robust mask-free formulation built on direct density prediction. We leverage the scene parameterization from mip-NeRF360~\cite{barron2022mip} to handle background clutter; by centering the hand in the parameterization's near-identity region, the network naturally isolates foreground geometry without explicit supervision. To recover high-fidelity surfaces from this density field, we integrate RefNeRF-inspired~\cite{verbin2022ref} appearance disentanglement, predicting view-independent diffuse color and view-dependent specularities to prevent baked-in lighting. Furthermore, we enforce surface sharpness and geometric consistency using density concentration~\cite{barron2022mip} and normal orientation priors~\cite{verbin2022ref}. This combination ensures the extraction of clean, coherent meshes (Fig.~\ref{fig:ablation_surface_detail}(e-f)) while maintaining the convergence stability of density fields.

\begin{figure}
    \centering
    \includegraphics[width=\linewidth]{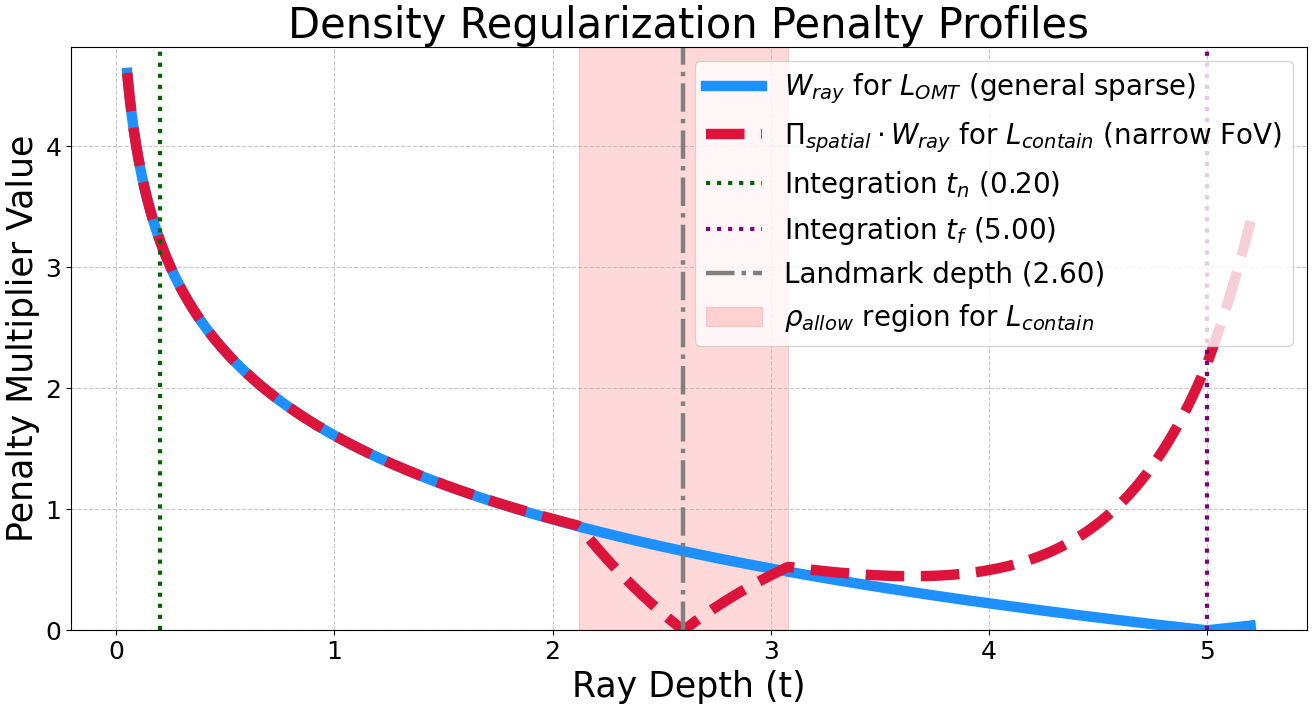}
    \caption{ Comparison of the penalty multiplier profiles for our two scenario-specific 
density regularization strategies. The x-axis represents depth along a ray; the y-axis shows the value of the 
penalty multiplier ($W_{{ray}}$ for $L_{{OMT}}$ or $\Pi_{{spatial}} \cdot W_{{ray}}$ 
for $L_{{contain}}$) that scales the predicted density. All penalties are defined 
with respect to integration bounds $t_n=0.2$ (green dotted line) 
and $t_f=5$ (purple dotted line).
The $L_{{OMT}}$ strategy, for general sparse captures, exhibits 
strong penalties near the integration bounds, effectively discouraging 
near-field artifacts.
The $L_{{contain}}$ strategy (red, dashed), for narrow-FoV captures, combines a 
near-field penalty, an exponential far-field penalty (with parameters 
$w_{{bg}}=1e-4$, $s=10$), and a spatial mask $\Pi_{{spatial}}$. 
The spatial mask's influence is centered around the landmark depth and within the $\rho_{{allow}}=0.5$ region 
(light red shaded area), focusing the penalty on regions distant from the landmark 
while also regularizing near ($w_{{near}}=1.0$) and far field densities.
}
    \label{fig:denisty_reg}
\end{figure}

\begin{figure}
    \centering
    \includegraphics[width=\linewidth]{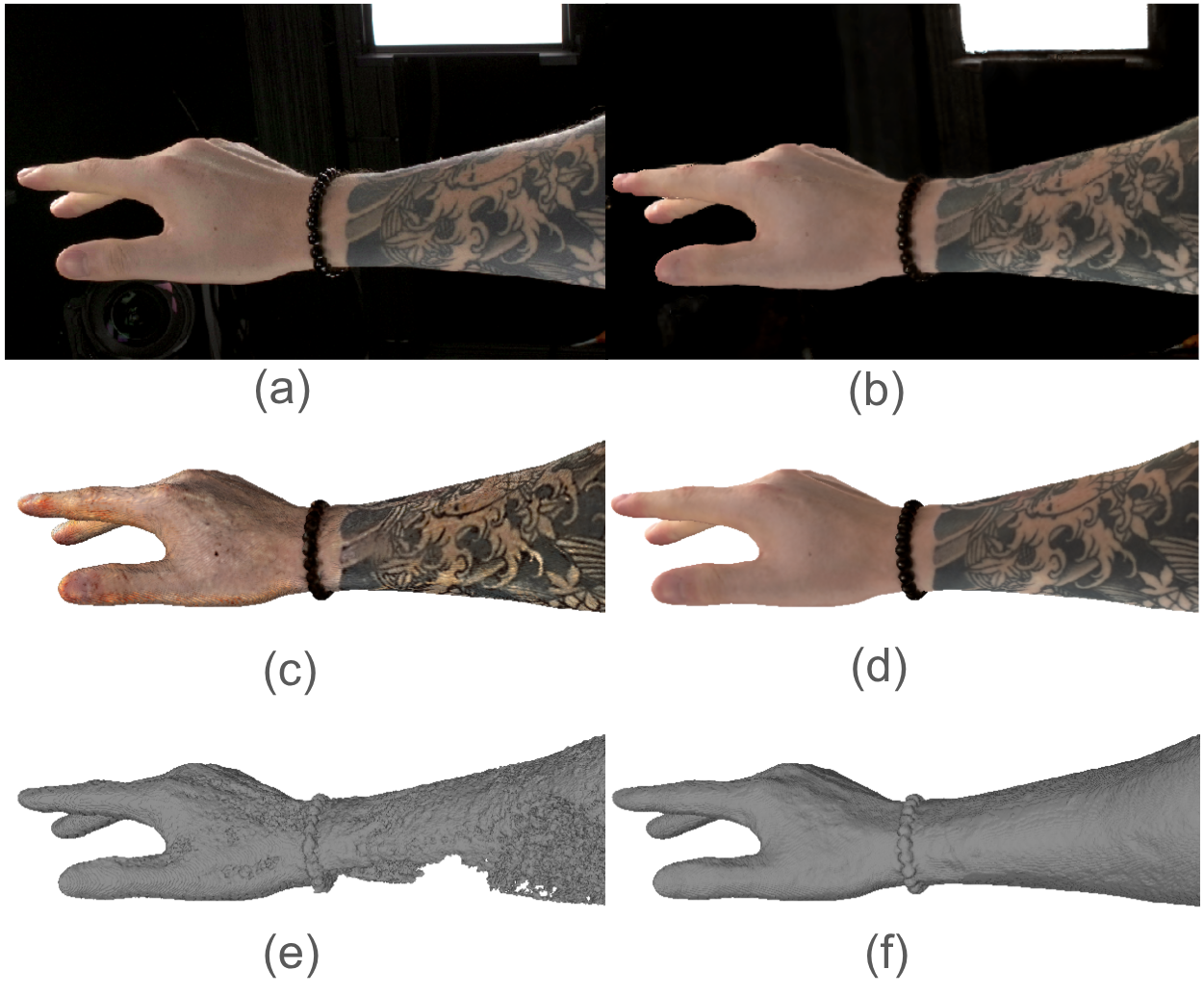}
    \caption{ Ablation study on surface detail enhancement and texturing. (a) Original captured image from one view. (b) Rendered diffuse color from our NeRF model, demonstrating significantly reduction on specular reflection (e.g.,upper arm). (c) Extracted texture using direct sampling at the reconstructed surface vertices, showing potential noise and inaccuracies. (d) Improved texture extracted using our proximity volume rendering technique (Sec.~\ref{sec:meshing}), which integrates diffuse color along surface normals for more robust and higher-fidelity appearance. (e) 3D reconstruction geometry without normal consistency and orientation losses (Sec.~\ref{sec:reconstruction_model}), exhibiting increased noise and less coherent surface details, highlighting the importance of these regularization terms. (f) Final 3D reconstruction geometry using our full pipeline, including all surface regularization losses.
}
    \label{fig:ablation_surface_detail}
\end{figure}

\begin{figure*}[htp]
    \centering
    \includegraphics[width=\linewidth]{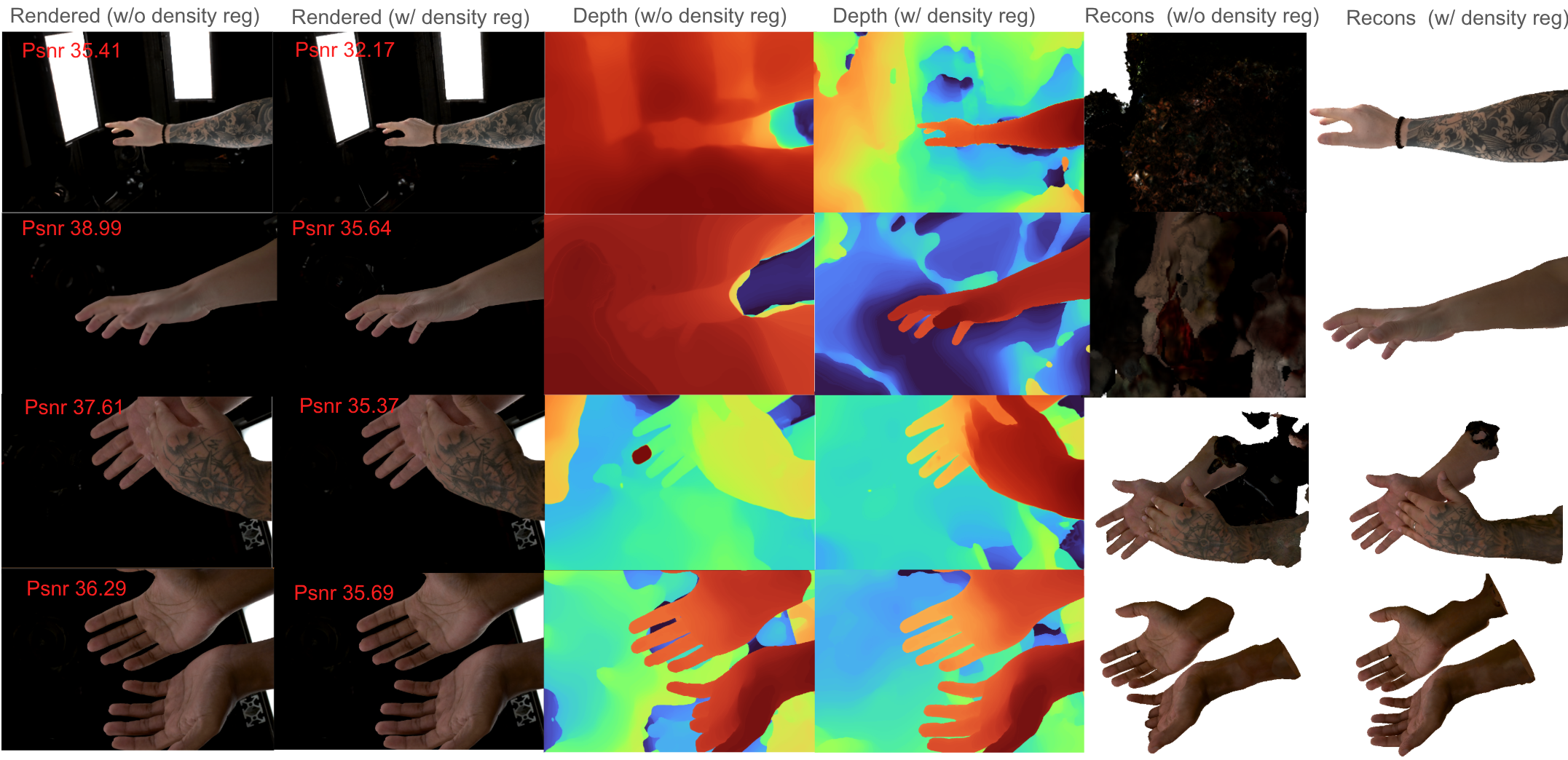}
    \caption{: Ablation study on the impact of our scenario-specific density regularization. Top two rows (wider FoV): OMT-inspired weighting ($L_{OMT}$) applied. Bottom two rows (narrow FoV): Density containment loss ($L_{contain}$) applied.
Our regularization significantly reduces artifacts like floaters and background collapse, leading to more coherent meshes. Note that while PSNR on rendered views might not always improve, this is because the regularizer is designed to improve geometric accuracy. It discourages the formation of geometrically implausible artifacts (like floaters or incorrect surfaces) even if such artifacts might coincidentally improve pixel-wise alignment with noisy or ambiguous parts of the input images.}
    \label{fig:unified_density_reg_ablation}
\end{figure*}

\subsection{Enhancing Geometric Robustness via Scenario-Specific Density Regularization}
\label{sec:density_regularization}

A prevalent challenge during NeRF optimization is the emergence of geometric artifacts, such as "floaters" (spurious density) and near-camera "hallucinations." Additionally, setup with sparser cameras often have limited view overlap, which can cause "background collapse", where foreground geometry is incorrectly placed far from the subject. While potentially yielding plausible renderings from certain views, these artifacts introduce incorrect geometry, hindering accurate surface extraction. To address these issues, we introduce a unified density regularization framework that can be customized for specific capture scenarios by adjusting its weighting components.

The general form of our density regularization loss, $L_{\text{density\_reg}}$, is:
$$
L_{\text{density\_reg}} = \int_{t_{n}}^{t_{f}} \Pi_{\text{spatial}}(\mathbf{r}(t)) \cdot \sigma(\mathbf{r}(t)) \cdot W_{\text{ray}}(t) dt
\label{eq:unified_density_reg}
$$
where $\sigma(\mathbf{r}(t))$ is the density at point $\mathbf{r}(t)$ along the ray. The term $\Pi_{\text{spatial}}(\mathbf{r}(t))$ is an optional spatial masking function that can emphasize or de-emphasize regularization effects in specific 3D regions. The term $W_{\text{ray}}(t)$ is a ray-dependent weighting function that modulates the penalty based on the depth $t$ along the ray. The specific definitions of $\Pi_{\text{spatial}}$ and $W_{\text{ray}}$ are chosen based on the capture setup and desired regularization effect.

We instantiate this framework for two primary scenarios:

\paragraph{Mitigating Near-Field Artifacts in General Sparse Captures.}
    For capture setups prioritizing large hand motions, often utilizing cameras with common or wider Fields of View (FoV), near-field floaters and hallucinations are a primary concern. In this scenario, we aim to suppress spurious density near the camera’s near plane ($t_n$) without overly constraining the far field. This is achieved by setting:
    
    \begin{itemize}
        \item $\Pi_{\text{spatial}}(\mathbf{r}(t)) = 1$ (no spatial mask).
        \item $W_{\text{ray}}(t) = N_{OMT} \cdot \left| \log\frac{t_{f}}{t} \right|$, where $N_{OMT} = \frac{1}{|\log(t_{f}/t_{n})|}$ is a normalization factor. 
    \end{itemize}

    The resulting loss inspired by Optimal Mass Transport, which we refer to as $L_{OMT}$, becomes:
    $$
    L_{\text{OMT}} = \frac{1}{|\log(t_{f}/t_{n})|} \int_{t_{n}}^{t_{f}} \sigma(\mathbf{r}(t)) \left| \log\frac{t_{f}}{t} \right| dt
    $$
    The logarithmic weighting term $|\log(t_f / t)|$ heavily penalizes density $\sigma(\mathbf{r}(t))$ close to $t_n$, effectively "pushing" artifacts away from the near field, while having less impact on the foreground object. The photometric loss remains the primary driver for reconstructing the hand density at the correct location. $L_{OMT}$ provides essential regularization for robust geometry in typical multi-view motion capture scenarios with practical camera counts.

 \paragraph{Enforcing Comprehensive Containment in Detail-Focused Narrow-FoV Captures.}
    When the primary goal is to capture fine hand appearance details, systems designed for view-efficiency (often using fewer cameras) may employ narrow FoV. This configuration significantly exacerbates geometric ambiguities due to limited view overlap and potentially uninformative backgrounds, increasing the risk of severe near-field artifacts and background collapse. For these challenging scenarios, we require stronger, more comprehensive regularization. Here, we define:
    
      $\Pi_{\text{spatial}}(\mathbf{r}(t))$ can be a spatially varying mask, for instance, guided by landmarks $\{\mathbf{l}_j\}$ via $\Pi_{\text{spatial}}(\mathbf{r}(t)) = \text{clip} ( \min_j \|\mathbf{r}(t) - \mathbf{l}_j\|_2 / \rho_{allow}, 0, 1 )$, which penalizes density far from the expected hand location. Alternatively, $\Pi_{\text{spatial}}(\mathbf{r}(t)) = 1$ if no explicit spatial mask is used.
      
     $W_{\text{ray}}(t)$ is designed to penalize density in both near and far fields, e.g., $W_{\text{ray}}(t) = w_{bg} \exp(s(t-t_n)) + w_{near} |\log(t_f/t)|$. The exponential term $w_{bg} \exp(s(t-t_n))$ discourages density accumulation deep into the background, while the $w_{near} |\log(t_f/t)|$ term (similar to $L_{OMT}$'s weighting) addresses near-field artifacts.
     
    This configuration, termed $L_{\text{contain}}$, actively penalizes density that resides far from an expected target region or too close to the camera. It acts as a crucial geometric prior to achieve plausible reconstructions in demanding narrow-FoV setups where view redundancy is inherently low.

In summary, our unified density regularization $L_{\text{density\_reg}}$ (Eq.~\ref{eq:unified_density_reg}), through scenario-specific choices for $\Pi_{\text{spatial}}(\mathbf{r}(t))$ and $W_{\text{ray}}(t)$, visualized in Fig.~\ref{fig:denisty_reg}, allows us to effectively address distinct geometric challenges. The OMT-inspired weighting is tailored for near-field artifact suppression in setups where near-camera geometry is under-constrained, while the more comprehensive containment weighting provides stronger constraints essential for robust reconstruction from challenging narrow-FoV data where achieving geometric consensus is difficult. Incorporating the appropriate instantiation of this regularization framework significantly improves geometric accuracy and the reliability of surface extraction from photometric data captured under such viewpoint limitations.

\subsection{Accelerating Training}
\label{sec:acc_train}

Making this high-fidelity optimization practical is paramount. Our primary acceleration comes from leveraging the multi-resolution hash grid encoding from Instant NGP [Müller et al. 2022]. This architecture drastically reduces computational cost, bringing the per-frame training time to approximately 20 minutes on an NVIDIA V100 GPU.
For further acceleration, particularly for large-scale batch processing, we introduce an optional segmentation-guided sampling strategy. We use a pre-trained segmentation network to predict a coarse hand region, which then guides our ray sampling to concentrate on the foreground. This focused sampling halves the training time to roughly 10 minutes per frame while also providing a marginal improvement in reconstruction quality. This two-tiered approach to efficiency, with hash grids providing the core speed-up and guided sampling offering an optional boost, was critical for processing the thousands of sequences in our dataset.

\begin{figure}
    \centering
    \includegraphics[width=\linewidth]{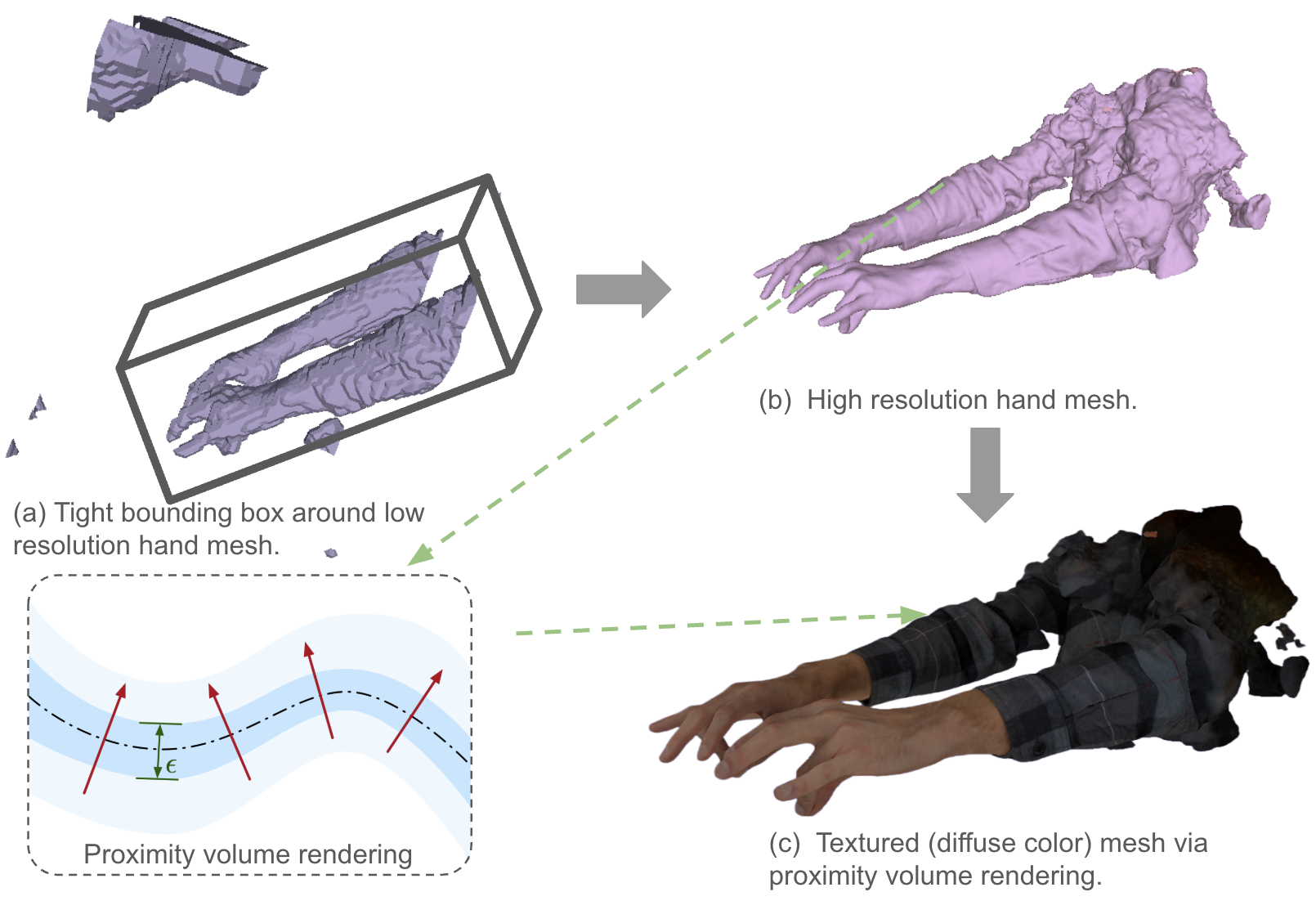}
    \caption{ Mesh extraction and texturing pipeline. (a) Initial mesh extraction via Marching Cubes within a defined region of interest (ROI), with a tight bounding box isolating the primary hand geometry from artifacts. (b) Resulting high-resolution hand mesh after refinement. (c) The final mesh textured with diffuse colors. Vertex colors are robustly determined by integrating the learned diffuse color field along surface normals in a local neighborhood, yielding high-fidelity appearance.}
    \label{fig:mesh_extraction_texturing}
\end{figure}

\subsection{Mesh Extraction and Texturing}
\label{sec:meshing}

To obtain an explicit 3D mesh for downstream applications, we extract a textured isosurface from the learned density field $\sigma(\mathbf{x})$. The process begins by applying the Marching Cubes algorithm~\cite{lorensen1998marching} at a density level $\tau$ within the foreground region of our scene parameterization. This yields an initial mesh, from which we select the largest connected component typically the hand to form $M_{hand}$. For higher detail, this process can optionally be repeated within a tighter bounding box around this initial result.

A key challenge is generating high-fidelity vertex colors that are robust to small inaccuracies in the extracted surface. Simply querying the learned diffuse color field $\mathbf{c}_d(\mathbf{x})$ at a vertex location $\mathbf{v}$ can be noisy, as the density field is not perfectly sharp at the isosurface. To address this, we introduce {\it proximity volume rendering}, a technique that computes a stable color estimate by integrating information in a small neighborhood along the surface normal $\mathbf{n}_v$. The final vertex color $\mathbf{c}_{vertex}(\mathbf{v})$ is computed as:
$$
\mathbf{c}_{vertex}(\mathbf{v}) = \frac{\int_{-\epsilon}^{\epsilon} T_{\mathbf{n}_v}(t) \sigma(\mathbf{v} + t\mathbf{n}_v) \mathbf{c}_d(\mathbf{v} + t\mathbf{n}_v) dt}{\int_{-\epsilon}^{\epsilon} T_{\mathbf{n}_v}(t) \sigma(\mathbf{v} + t\mathbf{n}_v) dt + \delta}
$$
where $T_{\mathbf{n}_v}(t)$ is the local transmittance along the normal, $\epsilon$ is a small integration range, and $\delta$ provides numerical stability. As shown qualitatively in Fig.~\ref{fig:ablation_surface_detail}(c, d), this local integration, weighted by density and transmittance, effectively averages out noise and produces a clean, high-fidelity appearance. For the purpose of registration, these vertex colors suffice.

\begin{figure*}
  \includegraphics[width=\textwidth]{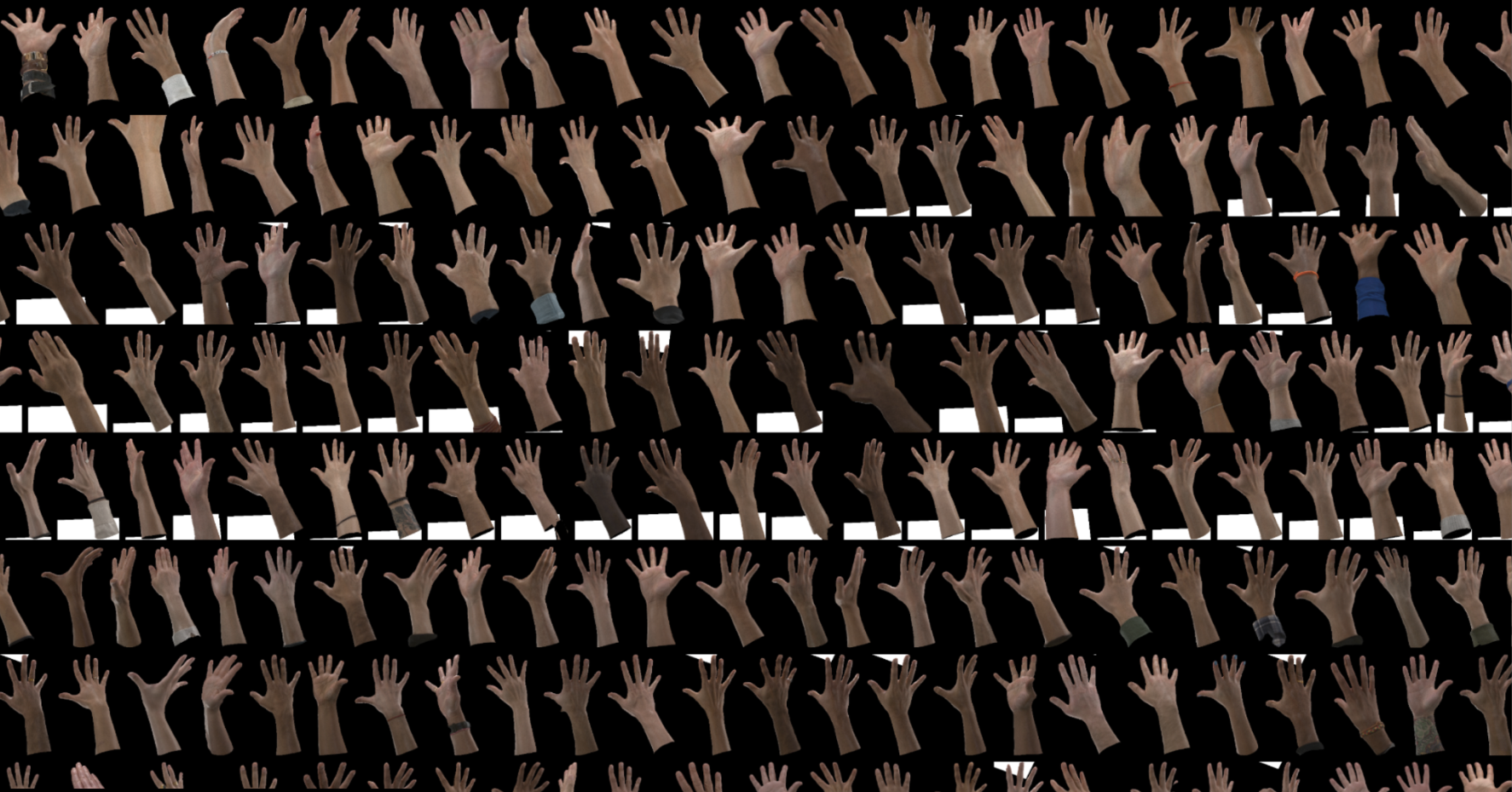}
  \caption{VEPHand enables scalable creation of high-fidelity, Personalized Hand Templates (PHT). This diverse sample, captured and processed by our end-to-end pipeline, showcases varied appearances. Underneath each textured surface lies a consistent intrinsic volumetric representation, empowering them to perform nuanced and complex hand articulations.}
  \label{fig:hand_appearance}
\end{figure*}

\section{Registration}
\label{sec:registration}

Generally, registration aims to find an optimal set of parameters, $\theta$, that govern a deformation model, $\text{Morph}$. This model transforms a source representation, $S$, to align with target data, $T$. The optimization typically minimizes an objective function that balances data fidelity against regularization priors:
\begin{equation*}
\theta^* = \text{argmin}_{\theta \in \Theta}  \text{Sim}(\text{Morph}(S; \theta), T) + \lambda_{\text{reg}} \text{Reg}(\theta) .
\label{eq:general_reg_abstract}
\end{equation*}
Here, the left term quantifies the similarity or alignment between the deformed source $\text{Morph}(S; \theta)$ and the target, while the right term represents regularization penalizing undesirable deformations. The weights $\lambda_{\text{reg}}$ balance these competing objectives, and $\Theta$ is the space of valid parameters $\theta$.

We instantiate this general framework with specific choices for each component. The following subsections will detail our approach:

     \textit{The Source Template ($S$):} We utilize a canonical, subject-specific hand template, which includes a consistent volumetric (tetrahedral) structure. Its automated generation for each subject is described in Suppl.~\ref{sec:PHT_generation}.
     
     \textit{The Deformation Model ($\text{Morph}$) and the Parameters ($\theta$):} Our Personalized Hand Model (PHM) detailed in Sec.~\ref{sec:deform_model}. This sophisticated model deforms the template $S$ using parameters $\theta$ (which include skeletal pose and fine-grained volumetric offsets) to produce the final posed shape, $\text{Morph}(S; \theta)$. It utilizes a consistent tetrahedral topology to manage volumetric deformations.
     
     \textit{Data Fidelity Terms ($\text{Sim}$):} We employ robust terms to measure alignment with various target data modalities derived from our reconstructions and input images (Sec.~\ref{sec:data_term}).
     
     \textit{Regularization Terms ($\text{Reg}$):} Comprehensive physics-inspired regularizers ensure plausible deformations and handle complex contacts (Sec.~\ref{sec:reg_terms}).
     
     \textit{Optimization and Initialization:} A multi-pronged initialization strategy is used to navigate the complex, high-dimensional optimization landscape robustly (Sec.~\ref{sec:initialization}).
     
     \textit{Handling Complex Scenarios:} The framework is extended to manage two-hand and hand-object interactions (Sec.~\ref{sec:extensions_two_hand_object}).

\begin{figure*}
    \centering
    \includegraphics[width=\linewidth]{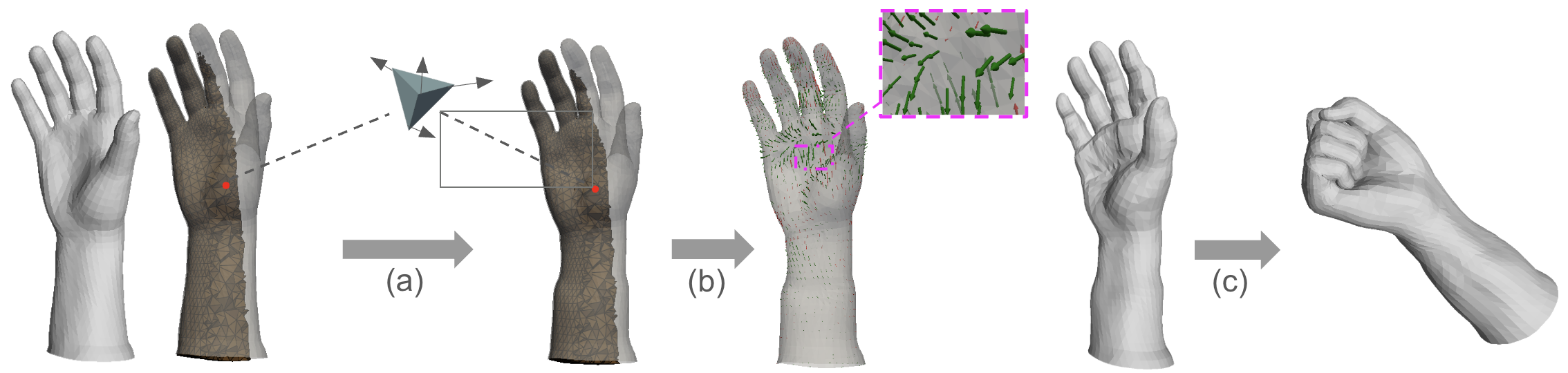}
    \caption{ Our Personalized Hand Model deformation: from intrinsic volume to posed surface. (a) A \textit{volume-based} intrinsic shape adjustment is first applied to the PHT's tetrahedral mesh via canonical offsets ($\Delta^c$). (b) This refined volume's surface then undergoes further \textit{surface-based} refinement using pose correctives ($P(\beta)$), visualized as vertex displacements (green arrows). (c) Standard \textit{surface-based} LBS, driven by pose parameters ($\beta$), generates the final articulated hand pose (e.g., fist).}
    \label{fig:hand_model}
\end{figure*}

\subsection{ A Personalized Dual-Component Volumetric Deformation Model}
\label{sec:deform_model}

Our registration is driven by a Personalized Volumetric Deformation Model, which deforms a subject-specific Personalized Hand Template (PHT) to match the input reconstruction. Each PHT, generated automatically, serves as a unique rest-state model containing a canonical surface mesh (vertices $v^c$), an underlying volumetric tetrahedral mesh (vertices $\rho^c$), a skeleton $J^c$, Linear Blend Skinning (LBS) weights $W$, and a set of pose-dependent corrective blendshapes $P(\beta)$. The surface mesh is an explicit boundary of the tetrahedral volume, ensuring that volumetric deformations coherently influence the surface.

The deformation itself is governed by two main parameter sets optimized per frame: skeletal pose parameters $\beta$ and frame-specific canonical tetrahedral offsets $\Delta^c$. Our key insight is to separate the deformation into complementary components, which are applied in a three-stage process (Fig. 9):
\begin{enumerate}[label=\arabic*), itemsep=0pt, topsep=1pt, parsep=1pt] 
    \item \textbf{Canonical Volumetric Deformation.} First, we model fine-grained, non-linear shape changes (e.g., muscle bulging) by applying intrinsic offsets $\Delta^c$ to the PHT's tetrahedral vertices ($\tilde{\rho}^c = \rho^c + \Delta^c$). By operating in the physically-grounded canonical space before articulation, we ensure these deformations remain consistent with the underlying volume.
    
    \item \textbf{Parametric Surface Refinement.} Next, we take the surface of the intrinsically deformed volume (vertices $\tilde{v}^c$) and apply the learned pose-dependent correctives $P(\beta)$ on it ($v^{c+P} = \tilde{v}^c + P(\beta)$). This step leverages the statistical prior of the base parametric model to compensate for low-frequency artifacts, such as volume loss at joints, that are characteristic of LBS.

    \item \textbf{Global Skeletal Articulation.} Finally, the k-$th$ refined canonical surface vertex $v^{c+P}_k$ is transformed into the target space via standard Linear Blend Skinning (LBS), driven by the skeletal parameters $\beta$:
    \begin{equation}
    v'_k = \sum_{j} W_{kj} K_j(J^c, \beta) (v^{c+P}_k).
    \label{eq:lbs_final_posing}
    \end{equation}

    Here $K_j(J^c, \beta)$ computes the rigid transformation matrix for joint $j$ based on the canonical joints $J^c$ and the current frame's pose parameters $\beta$.
\end{enumerate}

This dual-component framework combines the global articulation from LBS and the artifact correction from learned blendshapes with the fine-grained, physically-plausible detail captured by our intrinsic volumetric offsets. The parameters $\beta$ and $\Delta^c$ are optimized per frame to minimize an objective function balancing data fidelity (Sec.~\ref{sec:data_term}) and regularization (Sec.~\ref{sec:reg_terms}).

\subsection{Data Fidelity Terms} 
\label{sec:data_term}

Our registration pipeline aligns the deformable hand model with observations by minimizing a set of data fidelity terms. These terms are designed to be robust against the inherent challenges of hand capture, such as occlusions and reconstruction noise. We leverage both the 3D reconstructions from our neural pipeline and the original 2D multi-view images for a comprehensive alignment.

\paragraph{3D Landmark Alignment.}
The landmark term improves registration accuracy, particularly during initialization and for challenging poses, by aligning mesh landmarks with target landmarks, detailed in Suppl.~\ref{sec:landmark_estimation}, derived from 2D detections. We calculate mesh landmark positions as linear combinations of vertex coordinates and minimize the squared Euclidean distance between these positions and the target landmarks.

\paragraph{3D Surface Alignment.} To align our registered model, $M'$, with the target 3D reconstruction, $M^t$, we employ a robust point-to-plane distance metric (detailed in Suppl.~\ref{sec:data_term_details}). This term measures the geometric discrepancy between the two surfaces. To ensure reliability, we incorporate two key masking strategies: a normal compatibility mask prevents matching front-facing surfaces to back-facing ones, and a non-collision mask that down-weights the influence of vertices involved in self-contact. This results in a robust alignment that is resilient to outliers and geometrically implausible correspondences.

\paragraph{2D Photometric Consistency.} While 3D reconstructions provide strong geometric guidance, they can sometimes lack the fine textural detail present in the original images. To leverage this rich information, we introduce a photometric consistency term that operates in the 2D image space. Using a differentiable rasterizer, we render various attributes from our registered model $M'$ for each camera view, including color, normals, depth, and silhouette. These rendered images are then compared against their corresponding target attributes: the original RGB images and maps derived from the 3D reconstruction. To focus the comparison on reliable regions, we use per-pixel masks based on depth and normal consistency, ensuring the loss is only computed where the model and target are in close agreement. 

This dual-modality approach, combining 3D geometric alignment with 2D photometric refinement, allows our model to capture both the overall hand structure and its fine surface details accurately. The detailed formulations are provided in Suppl.~\ref{sec:data_term_details}.

\subsection{Physics-Inspired Regularization Terms} 
\label{sec:reg_terms} 

\begin{figure}
    \centering
    \includegraphics[width=\linewidth]{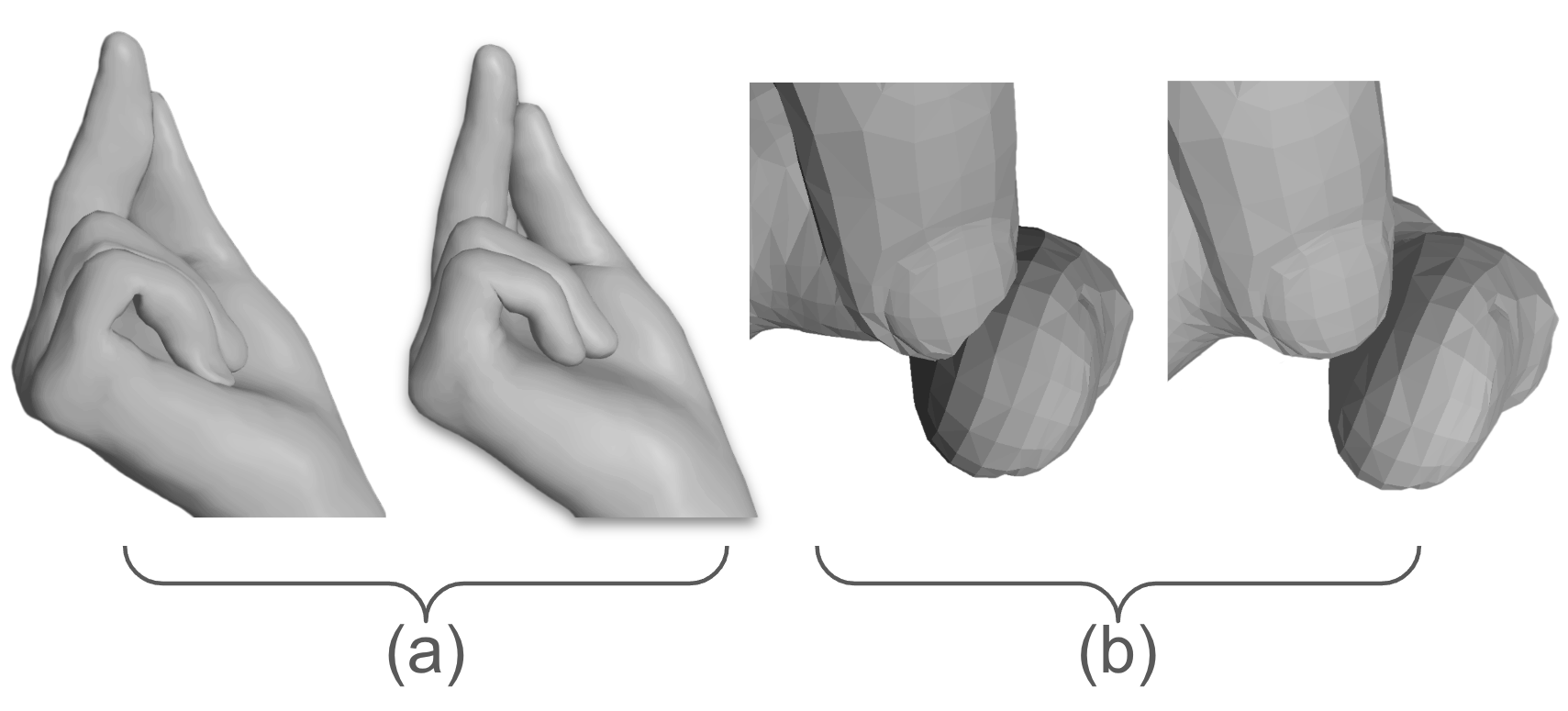}
    \caption{ Impact of physics-inspired regularization. (a): Without volumetric (tetrahedral) regularization, the surface can collapse unnaturally during extreme poses (left). Our volumetric model maintains plausible shape (right). (b): Without a collision penalty, surfaces interpenetrate (left). Our collision term effectively resolves penetration (right).}
    \label{fig:pyhsic_reg}
\end{figure}

To ensure that our registrations are physically plausible and robust to noisy data, we incorporate a comprehensive set of physics-inspired regularization terms. These terms prevent artifacts like self-intersections and unnatural deformations, guiding the optimization towards realistic hand poses and motions, particularly in cases of severe occlusion or complex articulation as shown in Fig.~\ref{fig:pyhsic_reg}.

\paragraph{Surface and Volumetric Shape Regularization.} We enforce plausible shape preservation for both the hand's surface and its internal volume. On the surface of the posed model $M'$, we use standard terms like an As-Rigid-As-Possible (ARAP) energy~\cite{sorkine2007rigid} to maintain local rigidity and a Laplacian smoothing term to encourage smoothness. Critically, to regularize the fine-grained intrinsic deformations ($\Delta^c$), we apply physics-inspired energies directly to the canonical tetrahedral mesh. This includes a volumetric ARAP term and a Neo-Hookean elastic model~\cite{bonet1997nonlinear}, which strongly penalizes volume changes and element inversions, ensuring the underlying tissue deformation behaves plausibly.

\paragraph{Collision Prevention.} Given the hand’s propensity for self-contact, a robust collision handler is essential. Our method first identifies vertices on the posed mesh that are inter-penetrating. For each, it finds opposing surface points on other parts of the hand while ignoring topologically adjacent regions. A penalty term is then applied to push these colliding surfaces apart, as demonstrated in Fig.~\ref{fig:pyhsic_reg}. This process also generates a mask used by our 3D data fidelity term (Suppl.~\ref{sec:collision_reg}) to reduce the influence of colliding vertices during alignment, preventing distorted fits.

\paragraph{Skeletal and Temporal Priors.} To maintain anatomical correctness and motion coherence, we apply two final regularization terms. A skeletal pose prior penalizes joint angles that deviate from a natural rest pose, preventing anatomically impossible configurations. For processing sequences, a temporal regularization term enforces smoothness by penalizing high-frequency changes in vertex positions, pose parameters, and the intrinsic volumetric offsets from one frame to the next.

Together, these regularization terms form a robust framework that ensures our final hand registrations are not only accurate with respect to the input data but also physically and anatomically plausible. Detailed formulations for all terms are provided in Suppl.~\ref{sec:reg_term_details}.

\subsection{Initialization and Optimization Strategy}
\label{sec:initialization}

Solving for the high-dimensional pose $\beta$ and volumetric offsets $\Delta^c$ constitutes a highly non-convex optimization problem, making a robust strategy for initialization and optimization essential.

\paragraph{Multi-Pronged Initialization.} To find a strong starting point for the per-frame optimization, we leverage a hybrid strategy that automatically selects the best of three different initial pose estimates. We evaluate a temporal prior from the previous frame's solution ($\beta(t-1)$), which provides stability for continuous motion. We also compute a landmark-based initialization by fitting the model to 3D landmarks detected in the current frame, offering responsiveness to abrupt movements. Additionally, after registering around 400 sequences, following TEMPEH~\cite{bolkart2023instant}, we train and employ a learned mesh prediction from a pre-trained multi-view network, which provides a strong, data-driven prior for complex articulations. By computing the initial objective score for all three candidates and proceeding with the best one, our system robustly handles a wide variety of motion without manual intervention. The canonical offsets $\Delta^c$ are always initialized from the previous frame to promote temporal smoothness.

\paragraph{Coarse-to-Fine Optimization Strategy.} We adopt a staged optimization approach to gently guide the solution. The process begins with a coarse 3D fitting stage, where we primarily optimize using the 3D data terms and the main shape regularizers, with the collision penalty either disabled or heavily down-weighted. This stage focuses on achieving a good overall alignment of the pose and shape. Following this, we proceed to a refinement stage, where we introduce the 2D photometric consistency term and fully activate the collision penalty. This allows the detailed image appearance to correct for subtle drifts and provides the hard constraints needed to resolve any remaining interpenetrations. The entire optimization is performed using the Adam optimizer.

\subsection{Extensions to Two-Hand and Hand-Object Scenarios}
\label{sec:extensions_two_hand_object}

Our registration framework can be readily extended to handle complex interactions, such as two-hand and hand-object scenarios, by employing a unified mesh concatenation strategy.

For two-hand registration, we instantiate separate models for the left and right hands. After deforming each individually, we concatenate them into a single mesh. All subsequent data fidelity and regularization terms are then applied to this combined entity. This elegantly allows our standard collision prevention mechanism to resolve both intra-hand and inter-hand penetrations simultaneously. To address potential forearm entanglement, a case where landmarks are missing, we add a brief refinement step that freezes the hands and optimizes only the forearm vertices with a stronger collision penalty, ensuring plausible separation.

Similarly, for hand-object interactions, we assume the object's geometry and pose are known (e.g., via marker tracking). We concatenate the posed hand mesh with the object mesh and apply our data and regularization terms. The collision prevention and 2D photometric consistency terms operate on the combined mesh, naturally handling hand-object interactions. The optimization, however, only updates the hand model's parameters ($\beta, \Delta^c$), as the object's state is fixed. This simple yet effective strategy allows us to robustly register hands in complex contact with their environment.

\begin{figure}
    \centering
    \includegraphics[width=\linewidth]{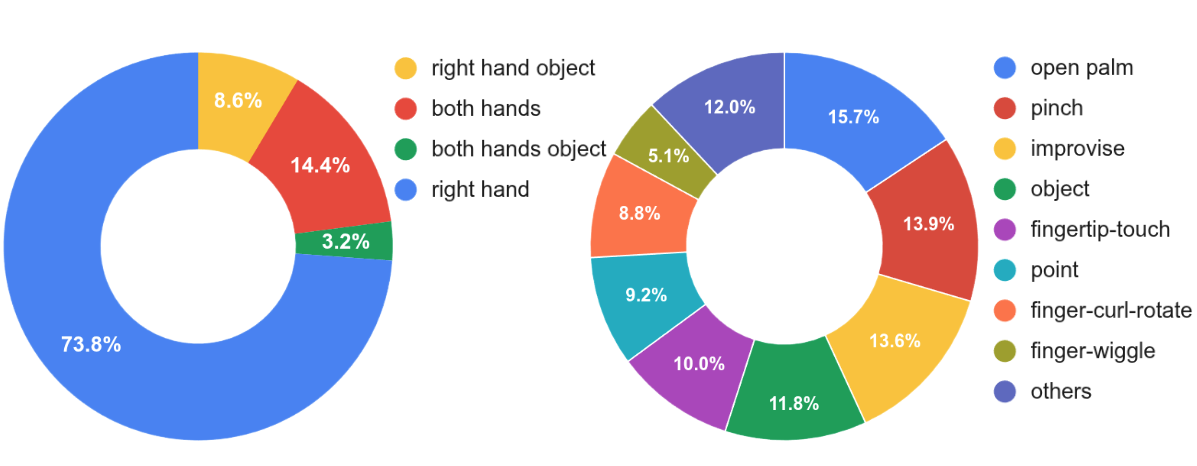}
\caption{Distribution of the Handbooth dataset across 11,781 sequences. (Left) Interaction scenario types. (Right) Captured hand pose/action types.}
\label{fig:dataset_distribution}
\end{figure}

\section{Results}
\label{sec:results}
\subsection{HandBooth Dataset}
\label{sec: handbooth_3d_dataset}
Our robust, automated pipeline enabled the creation of the \textit{Handbooth Dataset}, a large-scale collection of high-fidelity, dynamic hand performances. This dataset, featuring nearly 12,000 sequences from 800 subjects, serves as the primary validation of our system's scalability. Its scale and diversity, illustrated in Fig.~\ref{fig:dataset_distribution}, form the basis for valuable downstream assets. The outputs are organized into two primary forms: a core captured dataset of registered, textured meshes and a versatile synthetic dataset generated from it.

\paragraph{Handbooth Captured Dataset.} This is the direct output of our pipeline. It contains registered, textured 3D meshes for a diverse range of single-hand, two-hand, and hand-object interactions. The meshes capture rich geometric and textural details, including complex, pose-dependent phenomena like skin wrinkling and knuckle protrusions, which are often absent in purely parametric models. For object interactions, we provide precise alignment with known object geometries, tracked via ArUco markers. Furthermore, we process the captured appearance to recover high-quality PBR material maps~\cite{Weier2025PracticalInverse}, allowing for realistic relighting of the captured performances, with qualitative examples of the appearance quality shown in Fig.~\ref{fig:hand_appearance}.

\begin{figure}
    \centering
    \includegraphics[width=\linewidth]{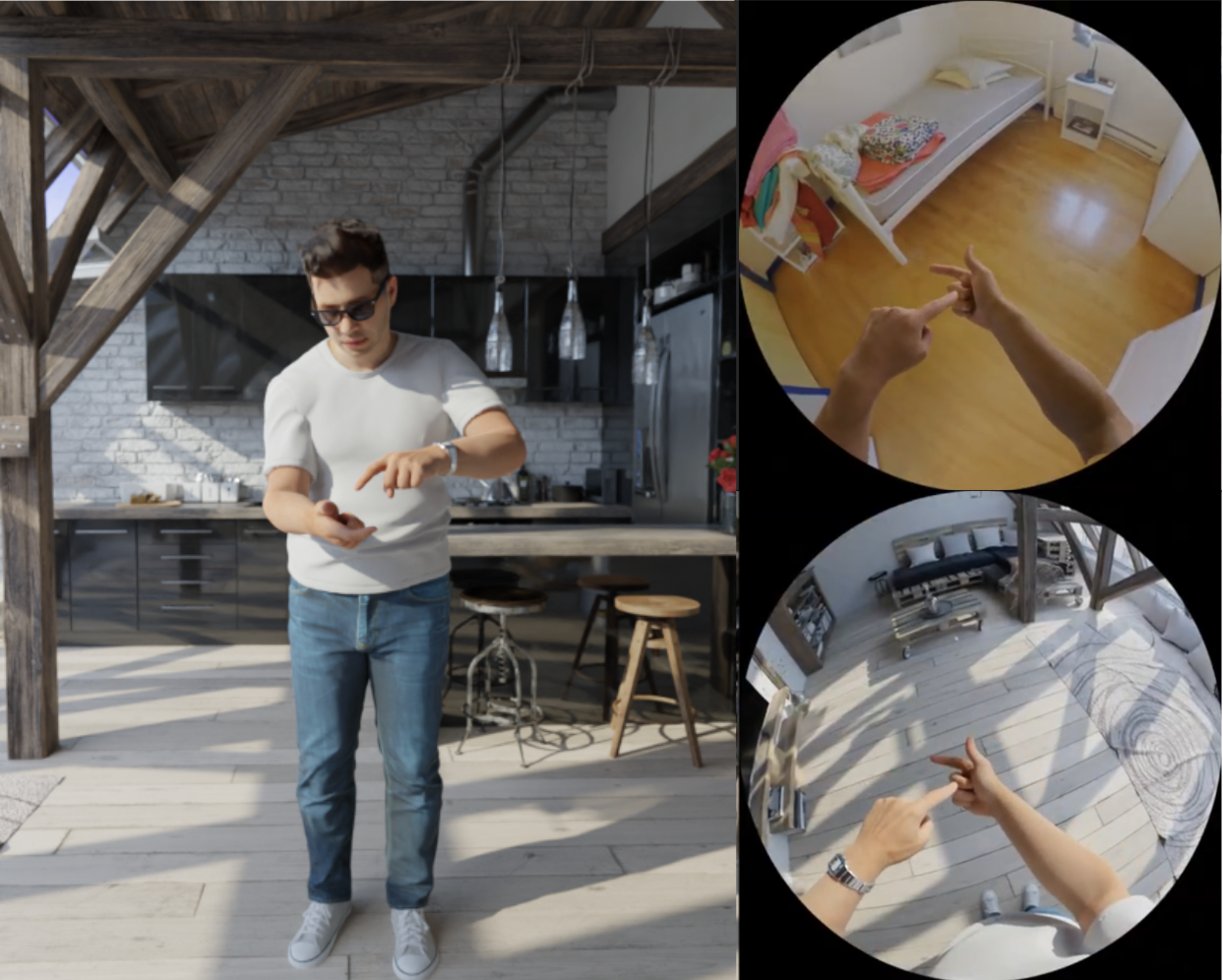}
\caption{Examples from our Handbooth synthetic dataset, generated by rendering our processed 3D hand performances. Left: A full-body parametric avatar with registered hand motion integrated into a synthetic scene. Right: Egocentric views simulating head-mounted display (HMD) perspectives, rendered with diverse backgrounds and illuminations. This dataset provides rich, varied, and accurately labeled data for training hand analysis models.}
\label{fig:body_synth}
\end{figure}

\paragraph{Handbooth Synthetic Dataset.} To facilitate the training of hand analysis models, we leverage our 3D assets to generate a large-scale synthetic dataset with perfect ground truth labels. To maximize diversity, we render the data using two complementary approaches:

\begin{enumerate}[label=\arabic*), itemsep=0pt, topsep=1pt, parsep=1pt] 

\item \textit{Direct Scan Rendering.} We render the captured 3D meshes directly. This approach, exemplified in Fig.~\ref{fig:teaser}, preserves the high-fidelity geometry and texture from our reconstructions, including accessories like watches or rings. While highly realistic, this method is limited to the captured forearm and cannot be reposed, which can lead to hands appearing to "float" in space.

\item \textit{Parametric Full-Body Rendering.} We use the registered hand motion to animate a full parametric body model, as shown in Fig.~\ref{fig:body_synth}. This places the hand in a naturalistic context, allowing for flexible camera framing and variations in body shape, skin tone, and clothing. The trade-off is a reduction in geometric and textural detail compared to the original scans.
\end{enumerate}

By rendering these assets in Blender with randomized camera poses (including egocentric HMD views), diverse lighting (HDRI), and varied backgrounds, we generate a rich and versatile dataset that couples realistic images with precise 2D and 3D labels.

\begin{figure}[htbp] 
    \centering 

    \begin{minipage}{0.9\linewidth}
    \centering
    \includegraphics[width=1.1\linewidth]{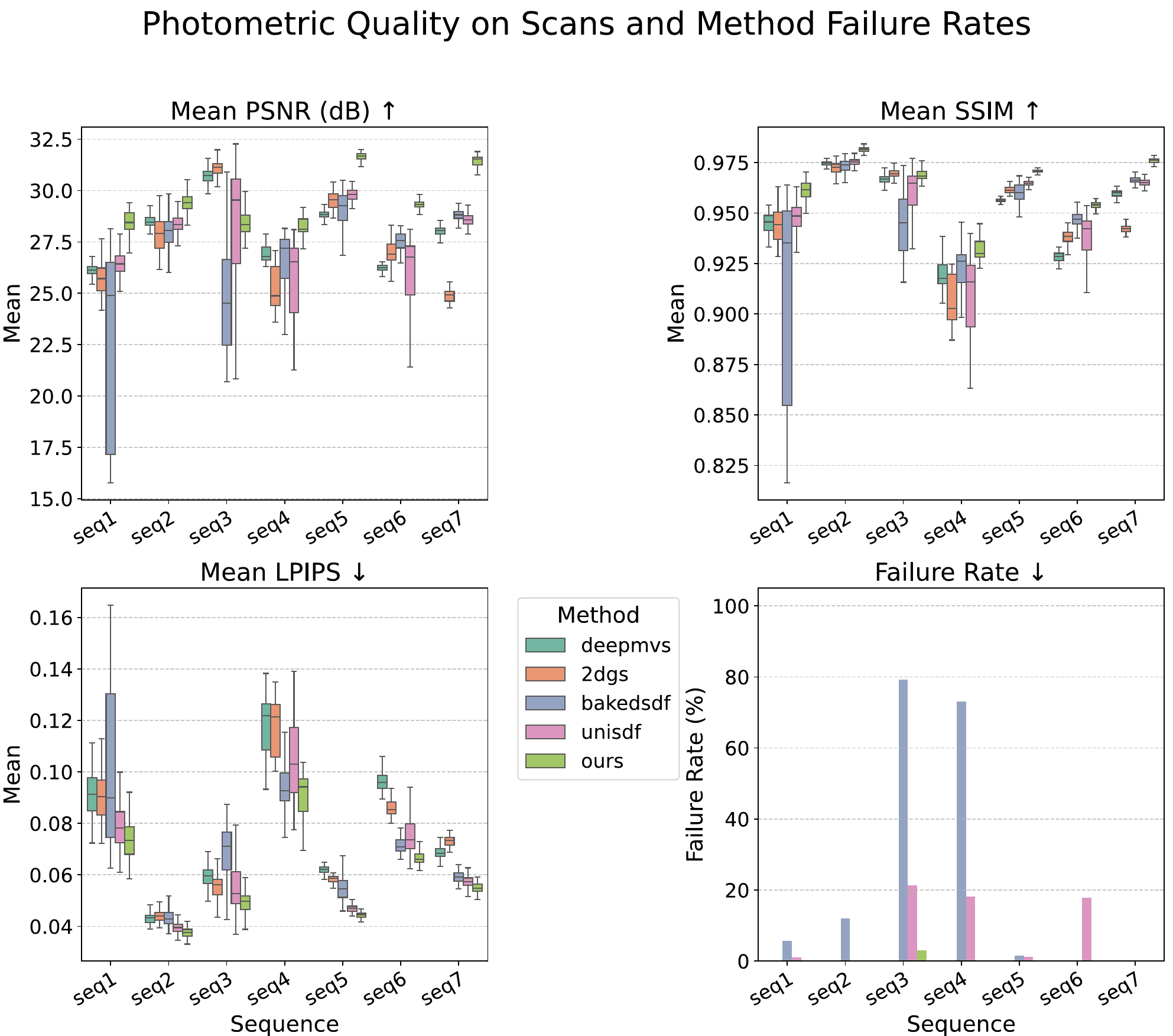}

    \label{fig:scan_quality_plot_part} 
\end{minipage}
    \vspace{0.5em} 

    \begin{minipage}{\linewidth} 
        \centering
        \scalebox{0.65}{%
            \begin{tabular}{l c c c c c} 
                \toprule
                \textbf{Method} & 
                \textbf{PSNR~$\uparrow$} & 
                \textbf{SSIM~$\uparrow$} &   
                \textbf{LPIPS~$\downarrow$} & 
                \textbf{Fail (\%)~$\downarrow$} &
                \textbf{T (min)~$\downarrow$} \\ 
                \midrule
                Colmap~\cite{schoenberger2016sfm, schoenberger2016mvs}   & 22.09 (1.862) & 0.90 (0.035) & 0.099 (0.0322) & N/A & 43 \\
                DeepMVS~\cite{qiu2024chosen}  & 27.99 (1.600) & 0.95 (0.020) & 0.075 (0.0265) & \textbf{0.0\%} & \textbf{5} \\
                2DGS~\cite{huang20242d}     & 27.35 (2.299) & 0.95 (0.024) & 0.075 (0.0262) & \textbf{0.0\%} & 9 \\
                \hdashline \noalign{\smallskip}
                Lang~\cite{li2023neuralangelo}     & 20.77 (1.087) & 0.90 (0.040) & 0.092 (0.0354) & 81.5\% & 17 \\
                BakedSDF~\cite{yariv2023bakedsdf} & 26.32 (3.668) & 0.94 (0.036) & 0.071 (0.0276) & 30.4\% & 26 \\
                UniSDF~\cite{wang2024unisdf}   & 27.46 (2.117) & 0.95 (0.025) & 0.066 (0.0240) & 8.9\% & 28 \\
                \midrule
                \textbf{Ours} & \textbf{29.26} (1.266) & \textbf{0.96} (0.017) & \textbf{0.061} (0.0196) & 0.5\% & 10 (20) \\ 
                \bottomrule
            \end{tabular}%
        } 
        \label{tab:overall_stats_compact_table_part} 
    \end{minipage}

    \caption{Quantitative comparison of reconstruction quality and robustness. (Top) Box plots showing the distribution of PSNR, SSIM, LPIPS, and Failure Rates across seven diverse dynamic hand sequences. (Bottom) Aggregate statistics. We compare VEPHand against Colmap and DeepMVS~\cite{qiu2024chosen}, 2D Gaussian Splatting (2DGS), and SDF-based neural methods (Lang~\cite{li2023neuralangelo}, BakedSDF~\cite{yariv2023bakedsdf}, UniSDF~\cite{wang2024unisdf}). Note: Colmap results are computed on rasterized point clouds as it fails to generate sufficiently dense geometry for meshing. For 2DGS, we used robust dense-ball initialization and TSDF-based mesh extraction. VEPHand achieves the best balance of photometric accuracy and robust convergence (0.5\% failure) in this sparse, unmasked setting. For all Nerf methods, we take NGP\cite{muller2022instant} as backbone. 10(20) refers to time cost w and w/o weighted sampling~\ref{sec:acc_train}.}
\label{fig:scan_quality_and_table}
\end{figure}

%

\begin{figure}
    \centering
    \includegraphics[width=\linewidth]{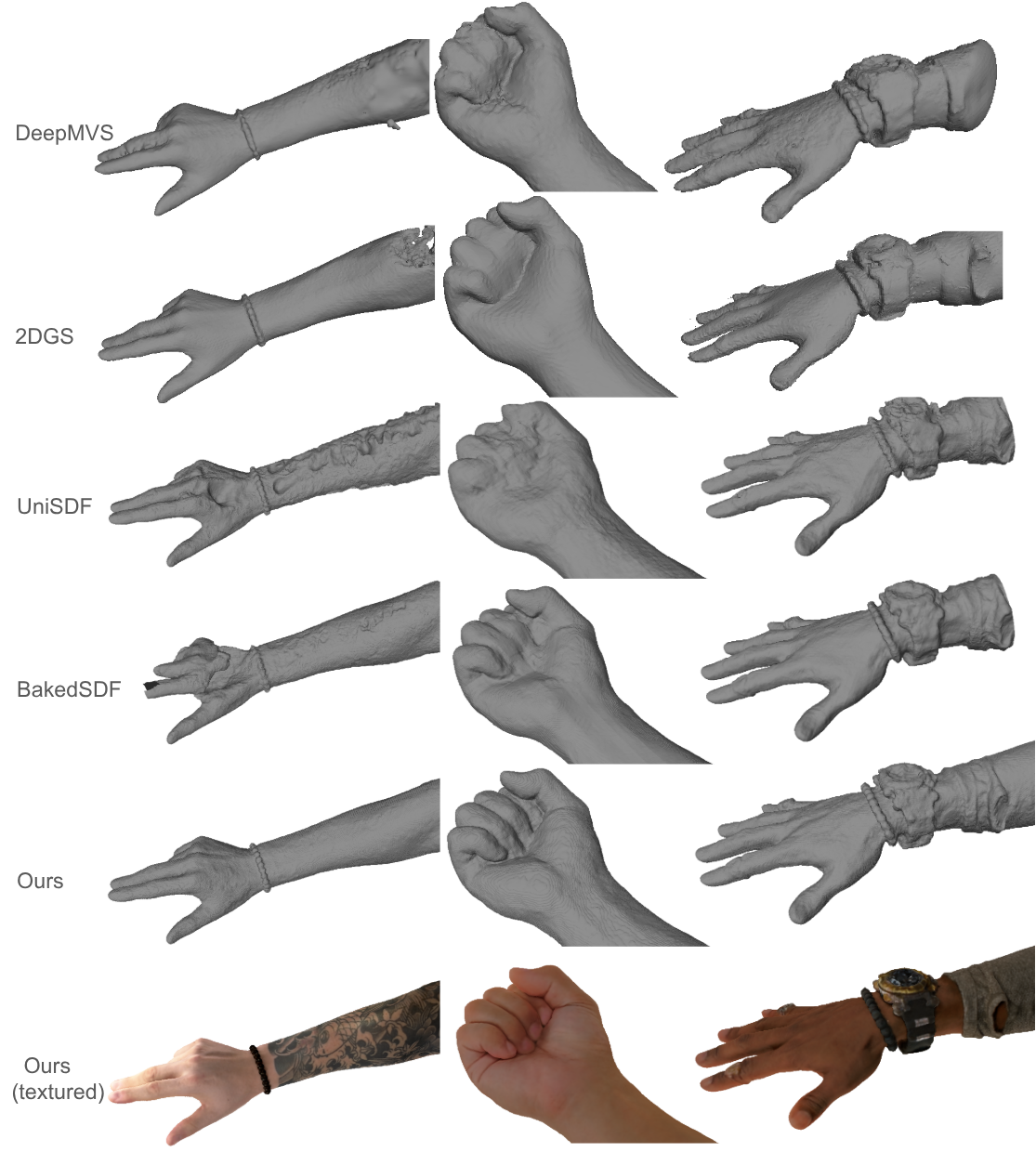}
    \caption{Qualitative comparison of hand reconstruction from sparse multi-view inputs. DeepMVS struggles with noise and artifacts in regions with limited view overlap, particularly in finger gaps. 2DGS yields characteristically over-smoothed surfaces that fail to capture fine anatomical details like knuckles. SDF-based methods (UniSDF, BakedSDF) are sensitive to the sparse setup, resulting in severe noise and topological holes. In contrast, our method produces  more complete and detailed geometry on skin wrinkles and complex accessories. Other methods like NeuS~\cite{wang2021neus} and Neuralangelo~\cite{li2023neuralangelo} are omitted as they failed to produce meaningful meshes from our sparse data.}
    \label{fig:nerf_compare_new}
\end{figure}
\begin{figure}
    \centering
    \includegraphics[width=\linewidth]{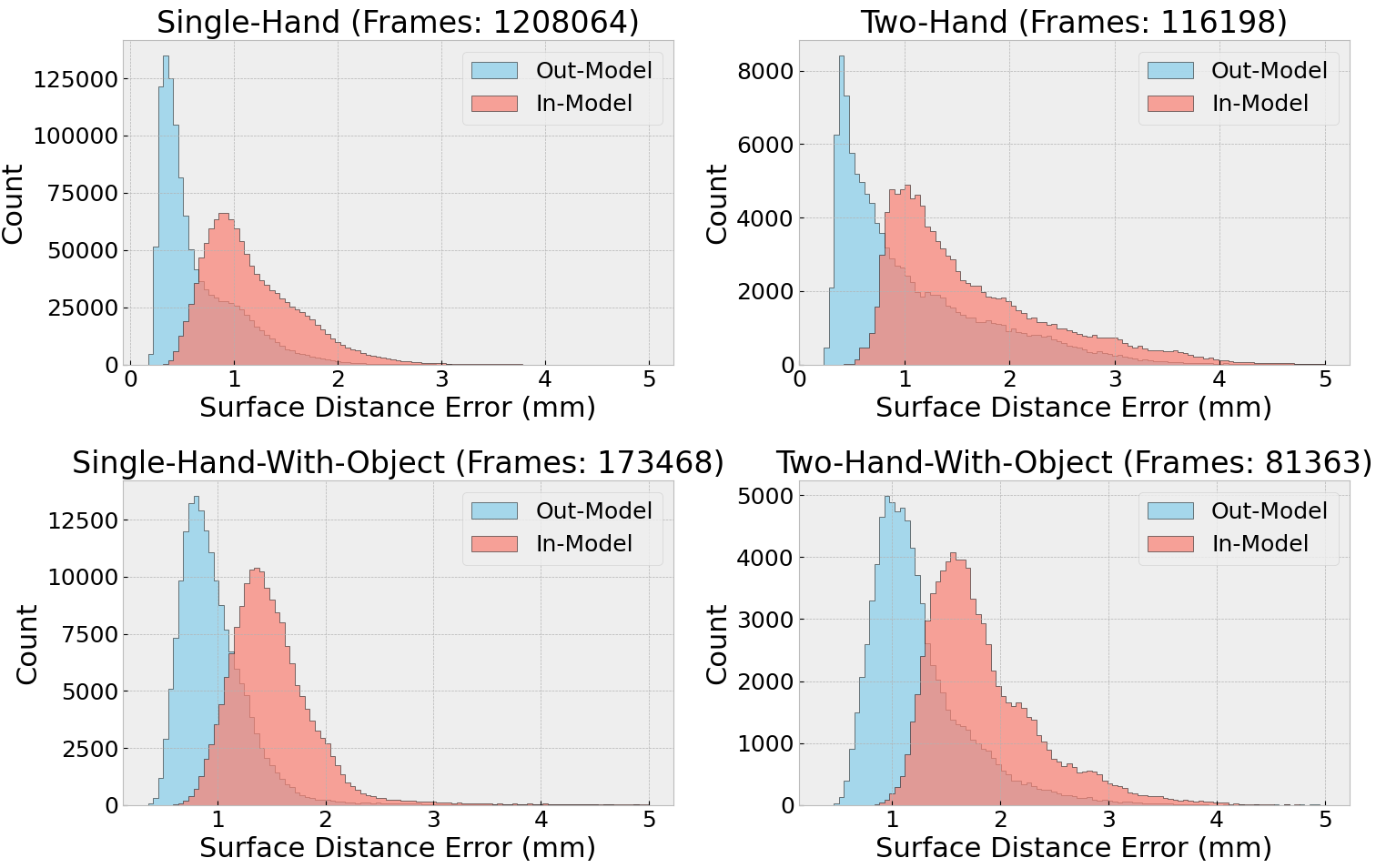}
    \caption{Registration accuracy evaluation via surface distance error (mm) histograms. Our full model including canonical tetrahedral offsets ($\Delta^c$ (Out-Model, blue) consistently outperforms a baseline without offsets (In-Model, red) across four interaction types (frame counts indicated). Lower errors demonstrate improved geometric fidelity.
}
    \label{fig:registration_errors}
\end{figure}
\begin{figure}
    \centering
    \includegraphics[width=\linewidth]{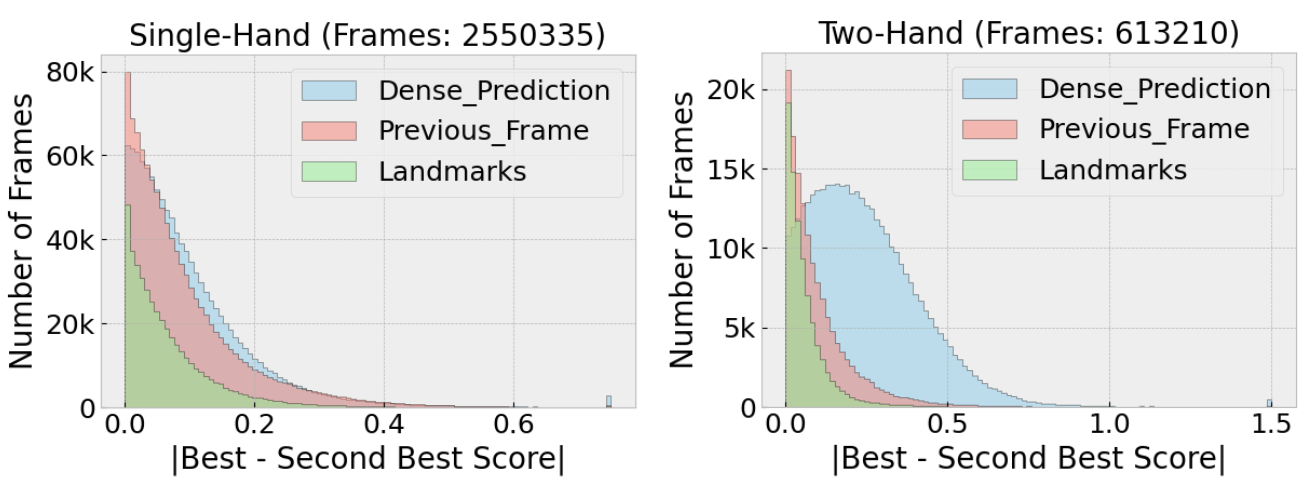}
\caption{Ablation study on registration initialization strategies. The histograms show the distribution of the absolute difference between the best (lowest) and second-best initial score (Chamfer distance in mm to the target scan) for single-hand (left) and two-hand (right) sequences. The color of each bar indicates which strategy yielded the best score for that frame: Learned Mesh Prediction (Dense\_Prediction, blue), Temporal (Previous\_Frame, red), or Landmark-based (Landmarks, green). The x-axis represents the magnitude of this score difference. Larger values indicate a clearer benefit from the chosen initialization strategy, validating our hybrid approach of evaluating all three and selecting the best seed for optimization.}
\label{fig:initialization_comparison}
\end{figure}

\subsection{Study on Reconstruction Quality}
\label{sec:study_recons}
We evaluated VEPHand's reconstruction against methods ranging from classical MVS, representative SDF Nerf approaches to Gaussian Splatting to validate its performance in our challenging view-efficient, unmasked setting.

\paragraph{Qualitative Comparison.} As illustrated in Fig.~\ref{fig:nerf_compare_new}, VEPHand consistently delivers complete and topologically sound geometry, whereas other paradigms exhibit distinct failure modes. For Multi-View Stereo (MVS), standard Colmap~\cite{schoenberger2016sfm, schoenberger2016mvs} failed to generate dense enough correspondences for meaningful mesh generation. To give MVS a fair chance, we trained a DeepMVS~\cite{qiu2024chosen} on 50,000 VEPHand reconstructions; this domain-specific adaptation significant improved the performance but the method still struggled in regions with limited view overlap, producing noisy surfaces that required aggressive Poisson smoothing to be usable. Similarly, for 2D Gaussian Splatting (2DGS)~\cite{huang20242d}, we tuned the parameters carefully and initialized the model as a dense volumetric ball to bypass the fragility of SfM point clouds in our setup. While this strategy yielded stable convergence, the resulting meshes (extracted via TSDF fusion) were characteristically over-smoothed, failing to capture fine-scale skin details.

In contrast, existing SDF-based methods demonstrated common difficulties in such unconstrained scenarios. Earlier approaches like NeuS~\cite{wang2021neus} and Neuralangelo~\cite{li2023neuralangelo} frequently suffered catastrophic failures; without accurate foreground masks or dense coverage, they often yielded collapsed geometry or failed to converge entirely. Even more recent methods designed for greater robustness, such as BakedSDF~\cite{yariv2023bakedsdf} and UniSDF~\cite{wang2024unisdf}, while showing improved stability, often produced noisy, incomplete results characterized by topological breaks (e.g., missing fingers) and a distinct lack of fine detail. These issues typically stem from their sensitivity to initialization or inherent difficulties in defining an explicit surface from ambiguous, view-efficient data. VEPHand, by avoiding these pitfalls, robustly handles background elements to recover coherent high-fidelity surfaces.

\paragraph{Quantitative Analysis.} These observations are validated by our quantitative analysis on seven diverse sequences, 3391 frames in total including single-hand, two-hand, hand-object captures, summarized in Fig.~\ref{fig:scan_quality_and_table}. To compute reconstruction metrics, we reparameterize the textures using all 20 camera views and project them back into each view for comparison. VEPHand achieves a failure rate of 0.5\%, a dramatic improvement over the SDF baselines (8.9\%--81.5\%). While DeepMVS and 2DGS also shows great stability, VEPHand outperformed them on photometric and perceptual metrics. We achieved the highest mean PSNR (29.26 dB) and SSIM (0.96), and the lowest LPIPS (0.061). The lower LPIPS score is particularly indicative of our method's ability to retain sharp, perceptual details that are lost in the smoother 2DGS and DeepMVS results.

\subsection{Study on Registration Quality}
\label{sec:study_reg}

{\bf Study on Registration Errors.}
We quantitatively evaluate registration accuracy by measuring surface-to-surface distance between our registered model and the input reconstruction (Fig.~\ref{fig:registration_errors}), excluding the forearm and regions deemed occluded in the reconstruction where geometric accuracy may be lower. This is performed across $>$ 1.5 million frames from diverse single-hand, two-hand, and hand-object interaction scenarios. We compare our full model ("Out-Model"), which optimizes pose ($\beta$) and canonical tetrahedral offsets ($\Delta^c$) with physics-inspired volumetric regularization, against a baseline ("In-Model") that omits these detailed $\Delta^c$ offsets.
Results clearly show our full "Out-Model" achieving significantly lower surface distance errors (typically $<$ 1mm, centered around 0.5-0.7mm) across all scenarios. This highlights the critical role of the optimizable canonical tetrahedral offsets ($\Delta^c$) in capturing fine-grained, non-linear surface details present in the high-fidelity reconstructions, which the LBS-based "In-Model" cannot fully represent. 
The consistent high performance across varied scenarios, including complex two-hand and hand-object interactions, underscores the robustness of our pipeline. 
While results are strong, quality depends on input reconstruction fidelity. Extreme self-occlusion remains a challenge, relying on model priors. Future work could explore tighter reconstruction-registration integration. In summary, our registration pipeline, particularly the volumetric deformation model, delivers high-fidelity alignment to complex reconstructed hand data.

\paragraph{Comparison on Public Datasets.}
To verify that our registration framework generalizes beyond our own capture system and reconstruction pipeline, we evaluated its representational capacity on two external, high-quality public scan datasets: the \textit{MANO testing set}~\cite{romero2022embodied} (50 scans from 6 subjects) and the \textit{DHM dataset}~\cite{moon2020deephandmesh} (1082 valid\footnote{Though 33k mentioned in dataset page, only 1082 frames have valid scans.} scans from a single subject).
We compared our model against MANO, NIMBLE~\cite{li2022nimble}, Handy~\cite{potamias2023handy}, and UHM~\cite{moon2024authentic} by fitting each model to the scans and measuring the surface distance.
As summarized in Tab.~\ref{tab:registration_comparison}, our model achieves the lowest mean error on both datasets (0.26mm on MANO and 0.50mm on DHM). This represents a significant improvement over parametric MANO model (0.94mm) and learning-based personalized model like UHM (0.75mm). These results indicate that the additional flexibility provided by our intrinsic volumetric offsets allows for a much tighter fit to complex hand geometries, validating the method's effectiveness on data independent of our own capture pipeline.

\paragraph{Ablation Study on Initialization Strategy.}
We evaluated our hybrid pose initialization by comparing three strategies: Temporal ($\beta(t-1)$), Landmark-based ($\beta_{\text{Ldm}}$), and Learned Mesh Prediction ($\beta_{\text{pred}}$), shown in Fig.~\ref{fig:initialization_comparison}.
For both single-hand (left) and two-hand (right) scenarios, the Learned Mesh Prediction frequently provides the best initialization, often by a significant margin (larger x-axis values), particularly in the more complex two-hand cases where its holistic multi-view processing excels. Temporal initialization demonstrates high reliability, often being selected as optimal, especially when differences are small. Landmark-based initialization also contributes uniquely in a subset of frames.
Crucially, no single strategy universally outperforms the others. The optimal choice varies, and the performance difference between the best and second-best can be substantial. This validates our approach of evaluating all three candidates and selecting the one that minimizes the initial objective function. This hybrid method effectively leverages the complementary strengths of each strategy, enhancing the robustness and quality of our registration pipeline by consistently choosing the most promising starting point for optimization.

\noindent\textbf{Implementation Details and Performance.}
We provide a comprehensive listing of all hyperparameters for both the neural reconstruction and volumetric registration stages, along with a detailed breakdown of computational runtime and hardware specifications, in the Supp.~\ref{sec:computation}.

\section{Application}
\label{sec:application}

\subsection{Parametric Hand Model}
\label{sec:parametric_model}

The high-fidelity registered meshes produced by our VEPHand pipeline serve as a rich data source for learning advanced parametric hand models. We leverage this data to train a compact statistical model, denoted as $\mathcal{H}(\alpha, \beta)$, where $\alpha$ represents identity-specific shape characteristics and $\beta$ defines the pose.

\paragraph{Model Training.} Following established methodologies similar to MANO~\cite{romero2022embodied}, we optimize the model's core components including LBS weights, joint regressor, and principal components for identity and pose-driven shape variations to best reconstruct our training corpus. We trained our model using a diverse subset of VEPHand, 2,440 scans from 122 subjects. The training objective minimizes the reconstruction error between the parametric model and our registration outputs; specific energy formulations and optimization details are provided in Suppl.~\ref{sec:build_parametric_model}.

\paragraph{Evaluation and Comparison.} To demonstrate the quality of our learned model, we compare its representational power against established baselines: MANO~\cite{romero2022embodied} and GHUM~\cite{xu2020ghum}. We evaluated reconstruction accuracy on two test sets: internal Held-Out set has a diverse set of 400 scans from 20 subjects excluded from training and the public MANO test set to test generalization.

For all comparisons, we employ a standard iterative optimization strategy to fit each model to the target scans. This procedure alternates between optimizing global pose ($\beta$) and identity shape parameters ($\alpha$) to minimize the point-to-plane distance between the model surface and the target scan.
Results are summarized in Table~\ref{tab:parametric_model_comparision}. Our model demonstrates superior representational capacity on both our internal held-out set and external MANO dataset.
\begin{table}[htbp]
    \centering
   
    \label{tab:combined_hand_tables}
    
    \begin{subtable}[t]{0.45\columnwidth}
        \centering
        \caption{Full Registration models.}
        \label{tab:registration_comparison}
        \hspace*{-0.6em}\scalebox{0.75}{%
            \begin{tabular}{l | c c}
                \toprule
                \multirow{2}{*}{\textbf{Model}} & \multicolumn{2}{c}{\textbf{Testing sets}} \\
                & \textbf{MANO} & \textbf{DHM} \\
                \midrule
                MANO~\cite{romero2022embodied}   & 0.94 & 1.36 \\
                NIMBLE~\cite{li2022nimble} & 0.88 & 1.22 \\
                Handy~\cite{potamias2023handy}  & 0.78 & 1.11 \\
                UHM~\cite{moon2024authentic}    & 0.75 & 0.59 \\
                \midrule
                \textbf{Ours}  & \textbf{0.26} & \textbf{0.50} \\
                \bottomrule
            \end{tabular}
        }
    \end{subtable}
    \hfill
    \begin{subtable}[t]{0.45\columnwidth}
        \centering
        \caption{Parametric models.} 
        \label{tab:parametric_model_comparision}
        \hspace*{-1.5em}\scalebox{0.75}{%
            \begin{tabular}{l c c}
                \toprule
                \multirow{2}{*}{\textbf{Model}} & \multicolumn{2}{c}{\textbf{Testing sets}} \\
                & \textbf{Ours} & \textbf{Mano} \\
                \midrule
                GHUM~\cite{xu2020ghum} & 1.01 & 1.12 \\
                MANO~\cite{romero2022embodied}      & 0.85 & 0.94 \\
                \midrule
                \textbf{Ours} & \textbf{0.69} & \textbf{0.76} \\
                \bottomrule
            \end{tabular}
        }
    \end{subtable}
     \caption{Quantitative evaluation of registration accuracy. (a) Comparison of our full registration framework against registration models on public scan datasets (MANO~\cite{romero2022embodied}, DHM~\cite{moon2020deephandmesh}). Our method, leveraging intrinsic volumetric offsets, achieves significantly tighter alignment. (b) Comparison of the representational power of our parametric model learnt from VEPHand scans against MANO and GHUM. Our model yields lower reconstruction errors (mm) on both an internal held-out set and the external MANO test set.}
\end{table}
\begin{figure} 
    \centering
    \includegraphics[width=\linewidth]{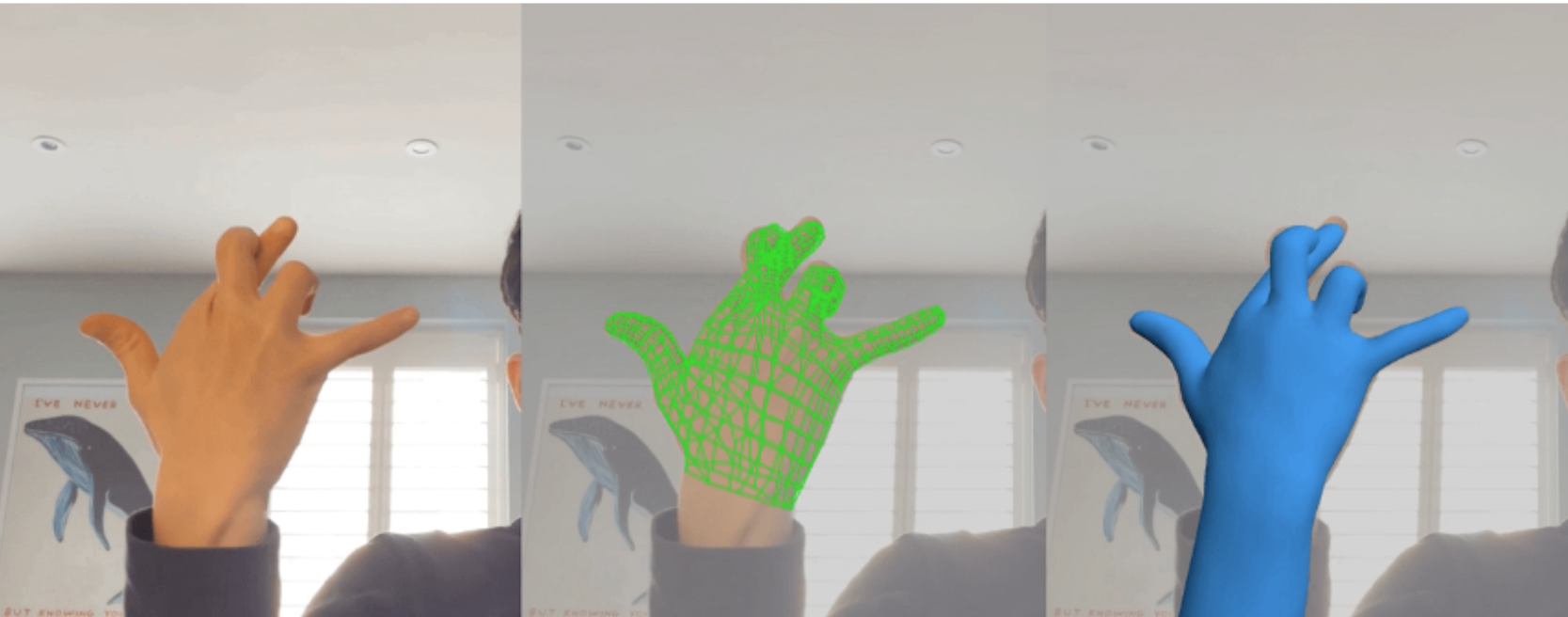} 
    \caption{Example output of our Dense Hand Tracker trained solely on the Handbooth 2D synthetic dataset. From left to right: input 2D image, predicted dense 2D landmarks overlaid (green), and the recovered parametric hand model fit (blue).}
    \label{fig:dense_hand_tracker_example}
\end{figure}

\subsection{Dense Hand Tracker}
Beyond generating parametric models, the extensive Handbooth 2D synthetic dataset (described in Sec.~\ref{sec: handbooth_3d_dataset}) also facilitates the training of robust deep learning models for 2D hand analysis. One such application we explored is the development of a Dense Hand Tracker. 
Similar to DenseLandMarks~\cite{wood20223d}, this tracker is trained to operate on single 2D images and simultaneously predict dense landmark outputs, a rich set of 2D landmarks distributed across the hand surface, as illustrated in Fig.~\ref{fig:dense_hand_tracker_example}). This provides a detailed, non-parametric representation of the hand's geometry and articulation directly from the image.

\section{Discussion}
\label{sec:discussion}
We introduce VEPHand, a robust, end-to-end pipeline for high-fidelity hand performance capture and registration using practical, view-efficient photometric systems. Our contributions address critical scalability and automation challenges through two primary innovations: mask-free neural reconstruction and physics-inspired volumetric registration.

The integration of these components into our "Handbooth" system and the subsequent processing of an extensive dataset of nearly 12,000 sequences underscore the scalability and robustness of VEPHand. We have shown its effectiveness across diverse scenarios, including single hands, intricate two-hand interactions, and detailed hand-object manipulations. The resulting high-quality textured meshes and registered parametric models provide a rich resource, as demonstrated by the creation of an improved parametric hand model and the training of a dense 2D hand tracker from our synthetic dataset. This work effectively bridges the gap between highly specialized, dense capture systems and more accessible but often lower-fidelity setups, paving the way for broader adoption of high-quality dynamic hand models.

\paragraph{ Limitations and Future Work.}
Despite these advancements, several limitations and avenues for future research remain:

\textbf{Reconstruction Fidelity.} While our reconstruction robustly handles captures with limited overlaps, experiments on the dense InterHand2.6M dataset revealed that our reconstructions were smoother than their reference scans. This difference is inherent to the algorithms: unlike active stereo or dense MVS systems which leverage projected patterns and explicit patch-matching to recover high-frequency surface detail, our density field optimization tends to regularize such variations to maintain topological consistency. However, this represents a strategic trade-off. By prioritizing view-efficiency, VEPHand achieves the flexibility to capture a range of poses and natural interactions comparable to dense systems but with significantly reduced hardware complexity, accepting a minor sacrifice in micro-surface detail for enhanced scalability.

\textbf{Pipeline Efficiency and 4D Representations.} The per-frame optimization, although accelerated by hash-grid encoding, remains a computational bottleneck (comprehensive parameters and computational performance analysis are provided in the Suppl.~\ref{sec:computation}). Future work could target amortized inference or transition to explicit dynamic representations like 4D Gaussian Splatting~\cite{idil2026grasp}. Given the high robustness we observed with 2DGS in our experiments, a temporal 4DGS with local geometry recovery shows significant potential. Such a method could allow for the direct regression of hand states, potentially bypassing the heavy volumetric reconstruction stage entirely for tasks that do not require explicit reconstruction, thereby simplifying the pipeline and approaching real-time performance.

\textbf{Deformable and Random Objects.} For hand-object interactions, we currently assume known object geometry and pose. Extending the system to handle unknown or articulated objects, potentially by co-segmenting and co-reconstructing them alongside the hands, similar to HOLD~\cite{fan2024hold}, but adapted to our unmasked, complex scenarios, would greatly enhance the system's versatility.

\paragraph{Conclusion.}
Overall, VEPHand represents a significant step forward in making high-fidelity, scalable hand performance capture more practical. By addressing critical bottlenecks in mask-free reconstruction and automated volumetric registration, our pipeline offers a robust solution for acquiring rich datasets of dynamic hand motion. We believe this work will facilitate further advancements in digital human creation, human-computer interaction, robotics, and biomechanics, providing the tools and data necessary to better understand and replicate the complexity of human hands.

\balance
\bibliographystyle{plainnat}
\bibliography{handbooth}

\clearpage

\appendix

\section{Robust 3D Landmark Estimation}
\label{sec:landmark_estimation}

Accurate 3D hand landmarks serve as valuable geometric constraints, particularly for registration initialization (Sec.~\ref{sec:initialization}) and potentially guiding reconstruction (Sec.~\ref{sec:density_regularization}). However, obtaining reliable 3D landmarks from multi-view images presents challenges, especially when using off-the-shelf 2D detectors trained primarily on frontal views. We observe that standard detectors like MediaPipe \cite{lugaresi2019mediapipe} can exhibit reduced accuracy on side or top-down views common in multi-view capture setups. Therefore, we employ a two-stage process involving robust triangulation followed by temporal refinement to compute reliable 3D landmarks $\{L_k^t\}$ (where $k$ indexes the landmark and $t$ the frame) from potentially noisy multi-view 2D predictions $\{l_{k}^{c,t}\}$ from camera $c$.

\paragraph{Stage 1: Per-Frame Robust Triangulation.}
For each frame $t$, we first apply MediaPipe independently to each camera view $c$ to obtain initial 2D landmark predictions $\{l_{k}^{c,t}\}$. To mitigate the impact of unreliable detections from certain views, we perform a robust triangulation using RANSAC \cite{fischler1981random} to estimate initial 3D landmark positions $L_k^{\text{init}, t}$. Specifically, for each landmark $k$:
\begin{equation}
    L_k^{\text{init}, t} = \text{RANSAC} \left( \text{Triangulate}, \{l_{k}^{c,t}\}_{c}, \{P_c\}_c \right)
    \label{eq:ransac_triangulation}
\end{equation}
where $\{P_c\}_c$ are the calibrated camera projection matrices. The RANSAC procedure involves repeatedly sampling minimal subsets of camera views, performing standard linear triangulation
to generate candidate 3D points $\{L_k^{\text{cand}}\}$, and evaluating the consensus of these candidates (e.g., based on reprojection error across all views). The candidate $L_k^{\text{cand}}$ with the largest consensus set (or derived from the largest consistent cluster of candidates) is chosen as the robust initial estimate $L_k^{\text{init}, t}$, effectively rejecting outliers from poor views. 

\paragraph{Stage 2: Temporal Refinement.}
The per-frame estimates $L_k^{\text{init}, t}$ may still contain temporal jitter and noise. We refine the entire sequence of 3D landmarks $\{L_k^t\}$ simultaneously by minimizing an energy function that balances data fidelity with temporal smoothness and skeletal constraints:
\begin{equation}
    \{L_k^t\}^* = \text{argmin}_{\{L_k^t\}} \sum_t \left( E_{\text{Data}}^t + \lambda_{\text{Temporal}} E_{\text{Temporal}}^t + \lambda_{\text{Bone}} E_{\text{Bone}}^t \right)
    \label{eq:landmark_refinement_energy}
\end{equation}
where $\lambda_{\text{Temporal}}$ and $\lambda_{\text{Bone}}$ are weighting coefficients. The terms are defined as:
\begin{enumerate}[label=\arabic*), itemsep=0pt, topsep=1pt, parsep=1pt] 
    \item \textbf{Data Term ($E_{\text{Data}}^t$):} Enforces agreement between the refined 3D landmarks and the initial 2D detections, weighted by visibility:
    \begin{equation}
        E_{\text{Data}}^t = \sum_{k, c} w_{k,c}^t \cdot \rho \left( \| P_c(L_k^t) - l_{k}^{c,t} \|_2 \right)
        \label{eq:landmark_data_term}
    \end{equation}
    Here, $P_c(\cdot)$ is the projection operation for camera $c$, $\rho(\cdot)$ is a robust norm (e.g., Huber) to handle outlier detections, and $w_{k,c}^t$ is a visibility score. This score estimates the visibility of landmark $k$ in view $c$ at time $t$, down-weighting contributions from potentially occluded keypoints. It can be computed by fitting a template hand model to $L_k^{\text{init}, t}$, rasterizing the model in view $c$, and measuring the visible portion of a predefined region around the keypoint projection.

    \item \textbf{Temporal Term ($E_{\text{Temporal}}^t$):} Encourages smooth landmark trajectories by penalizing high velocities and accelerations:
    \begin{equation}
        E_{\text{Temporal}}^t = \sum_k \left( \| L_k^t - L_k^{t-1} \|_2^2 + \| (L_k^t - 2L_k^{t-1} + L_k^{t-2}) \|_2^2 \right)
        \label{eq:landmark_temporal_term}
    \end{equation}

    \item \textbf{Bone Term ($E_{\text{Bone}}^t$):} Enforces skeletal plausibility by penalizing changes in bone length between adjacent frames:
    \begin{equation}
        E_{\text{Bone}}^t = \sum_{(k, j) \in \text{Bones}} \left( \| L_k^t - L_j^t \|_2 - \| L_k^{t-1} - L_j^{t-1} \|_2 \right)^2
        \label{eq:landmark_bone_term}
    \end{equation}
    where $(k, j)$ represents pairs of landmarks connected by a bone segment in the hand's kinematic skeleton.
\end{enumerate}
Minimizing the objective function in Eq.~\ref{eq:landmark_refinement_energy} (e.g., using iterative least squares or gradient descent) yields the final temporally coherent and geometrically plausible 3D landmark sequence $\{L_k^t\}^*$, suitable for use in downstream tasks.

\begin{figure}
    \centering
    \includegraphics[width=\linewidth]{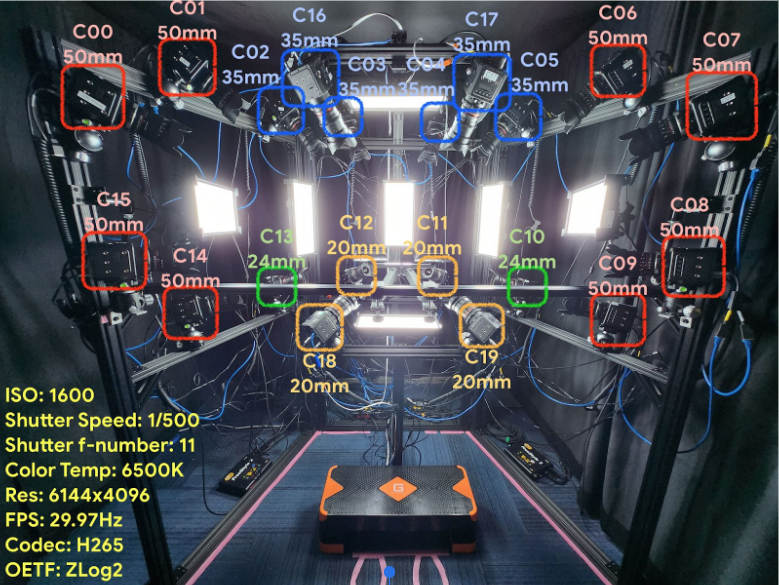}
    \caption{The Handbooth's narrow Field-of-View (FoV) configuration, designed for high-resolution texture and detail capture. Most cameras are equipped with longer focal length lenses ( (e.g., 20mm, 24mm, 35mm, and 50mm), color-coded by type) to maximize pixel density on the hand. This setup, while providing rich detail, presents geometric challenges due to reduced view overlap, addressed by specialized regularization in our reconstruction pipeline (e.g., $L_{\text{contain}}$). Key capture parameters are overlaid.}
    \label{fig:handbooth_small_fov_setup}
\end{figure}

\section{Enhanced NeRF Training with Mask-Guided Weighted Sampling}
\label{sec:fast_nerf}

Standard NeRF-based reconstruction pipeline treats the foreground (hand region) and the background agnostically. During the NeRF training stage, all the pixels from the images are sampled equally, of their relevance to the target object. Consequently, achieving high-quality reconstruction of a specific object, like a hand mesh, often requires extensive training iterations.

To accelerate the training speed, we introduce a mask-guided weighted sampling strategy. This approach leverages segmentation masks, even if they are not perfectly accurate, to focus the NeRF training process on the region of interest. The core idea is simple: concentrate sampling efforts on the hand region because our primary goal is to reconstruct the hand mesh, not the background.
The key steps of the proposed method is as follows:

\paragraph{Step 1: Segmentation Mask Generation.} For each captured view, a segmentation model generates a binary hand mask $M$, where pixel values are either 255 (hand) or 0 (background).

\begin{figure}[h]
    \centering
    \includegraphics[width=\linewidth]{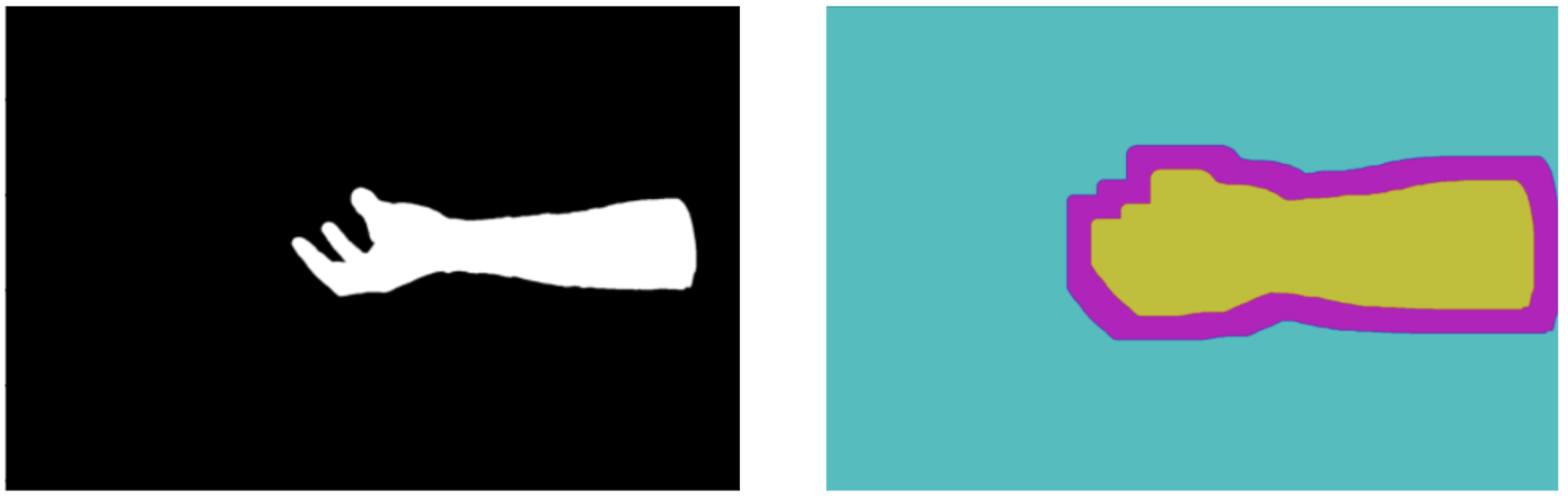}
    \caption{Pixel grouping calculated from a hand mask. Left side: the hand mask. Right side: grouped pixels, \textcolor{lime}{Group A}, \textcolor{magenta}{Group B} and \textcolor{cyan}{Group C}.}
    \label{fig:segmentation_groups}
\end{figure}

\paragraph{Step 2: Pixel Grouping via Dilation.} To create a prioritized sampling strategy around the hand, we perform sequential dilations on the mask $M$ and obtain three groups of pixels shown in~\ref{fig:segmentation_groups} as follows:

\begin{enumerate}[label=\arabic*), itemsep=0pt, topsep=1pt, parsep=1pt] 
    \item Group A: The mask  $M$  is first dilated 20 times using a $5\times5$ filter, resulting in $M'$. Group A consists of pixels where $M'\ge 128$. This group represents the core hand region and its immediate surroundings. 
    
    \item Group B: The dilated mask $M'$ is then dilated an additional 25 times, creating $M''$. Group B includes pixels where $M'<128$ and $M''\ge 128$. This group forms a transition zone around Group A.
 
    \item Group C: All remaining pixels constitute Group C, representing the background.
    
\end{enumerate}

\paragraph{Step 3: Weighted Sampling.} 
With the pixels being grouped into three different groups, different weights can be put on the different groups during the sampling. In detail, say $N$ pixels are to be sampled from the image space during one iteration of the training, then $p_A\times N$, $p_B\times N$ and $p_C\times N$ pixels will be sampled from Group A, B and C, respectively, where $p_A+p_B+p_C =1$. During the experiments, the sampling ratio for each group are set as $p_A$=0.5, $p_B$=0.1, and $p_C$=0.4, empirically.

\section{Build Parametric Hand Model}
\label{sec:build_parametric_model}

Given a set of $N$ registered hand meshes $\mathcal{M}^{\text{Reg}_i}$ (where $i$ indexes the frame/scan) from $S$ subjects, obtained from our registration pipeline (Sec.~\ref{sec:registration}), we can train a parametric hand model. This model, denoted as $\mathcal{H}(\alpha, \beta)$ where $\alpha$ are identity parameters and $\beta$ are pose parameters, aims to compactly represent hand shape and articulation. Our training process is inspired by MANO \cite{romero2022embodied}.

We start from a base rigged hand model that shares the same surface mesh connectivity as our registered hand meshes $\mathcal{M}^{\text{Reg}_i}$. This base model provides initial Linear Blend Skinning (LBS) weights $\tilde{W}$ and an initial kinematic skeleton with template joint locations $\tilde{J}^c_{\text{template}}$ (corresponding to the base generic model used for PHT generation in Sec.~\ref{sec:PHT_generation}).

The training minimizes the difference between the output of our parametric model and the registered hand meshes. We optimize for:
     $\bar{H}^c$: The mean hand shape (a set of template vertex positions),
     $W$: The final LBS weights,
     $\mathbf{R}_J$: A linear joint regressor that predicts 3D joint locations from a given hand shape,
     $B_s$: Per-subject identity-dependent shape offsets relative to $\bar{H}^c$,
     $P_i$: Per-frame pose-dependent corrective shape offsets,
     $\beta_i$: Per-frame pose parameters (e.g., joint rotations).

The optimization problem is formulated with the following energy term:
\begin{equation}
\begin{split}
E &= E_{\text{Data}} + E_{\text{Reg}} \\
E_{\text{Data}} &= \frac{1}{N}\sum_{i} ||\mathcal{M}^{\text{Reg}_i} - \mathcal{M}_{\text{model},i}||_2 \\
                & + \lambda_{\text{edge}} \frac{1}{N} \sum_{i} ||L(\mathcal{M}^{\text{Reg}_i}) - L(\mathcal{M}_{\text{model},i})||_2 \\
E_{\text{Reg}} &= \lambda_{\text{skinning}}||W - \tilde{W}||_2 \\
                &+ \lambda_{\text{joint}}||\mathbf{R}_J(\bar{H}^c) - \tilde{J}^c_{\text{template}}||_2 \\ 
                &+ \lambda_{\text{pose}}\frac{1}{N}\sum_{i}||P_i||_F^2 \\ 
                &+ \lambda_{\text{identity}}\frac{1}{S}\sum_{s}||B_{s}||_F^2 
\end{split}
\label{eq:parametric_model_energy}
\end{equation}
subject to $w_{kj}\ge0$ for all vertices $k$ and joints $j$, and $\sum_j w_{kj} = 1$.
Here, $L$ is the discrete Laplace-Beltrami operator. $\mathcal{M}_{\text{model},i}$ is the hand mesh generated by the model for frame $i$:
\begin{equation}
\mathcal{M}_{\text{model},i} = \text{LBS}(\bar{H}^c + B_{s(i)} + P_i, \mathbf{R}_J(\bar{H}^c + B_{s(i)}), W, \beta_i)
\label{eq:model_instance_training}
\end{equation}
where $s(i)$ denotes the subject corresponding to frame $i$.

We use the Adam optimizer to find the optimal $\bar{H}^c, W, \mathbf{R}_J, \{B_s\}, \{P_i\}, \{\beta_i\}$.
After optimization:
1.  We apply Principal Component Analysis (PCA) to the per-subject identity offsets $\{B_s\}$ to obtain a low-dimensional linear identity space, represented by a basis $\mathbf{B}$ and corresponding identity parameters $\alpha_s$. The mean identity shape is $\bar{H}^c$.
2.  We learn a linear pose corrective regressor $\mathbf{R}_{p}$ by solving the least-squares system $\mathbf{R}_{p} X_{\beta} = X_{P}$, where $X_{\beta}$ is a matrix of pose parameters $\beta_i$ (potentially transformed, e.g., into rotation matrices or their differences from rest) and $X_{P}$ is a matrix of the optimized pose correctives $P_i$.

The final parametric hand model $\mathcal{H}(\alpha, \beta)$ takes low-dimensional identity parameters $\alpha$ and pose parameters $\beta$ as input, and is defined as:
\begin{equation}
\mathcal{H}(\alpha, \beta) = \text{LBS}(H_{\text{canon}}(\alpha, \beta), J_{\text{canon}}(\alpha), W, \beta)
\label{eq:final_parametric_model_def}
\end{equation}
where $H_{\text{canon}}(\alpha, \beta) = \bar{H}^c + \mathbf{B}\alpha + \mathbf{R}_{p}(\beta)$ is the canonical hand shape including identity and pose-driven correctives (note that $\mathbf{R}_{p}(\beta)$ denotes applying the regressor to pose $\beta$), and $J_{\text{canon}}(\alpha) = \mathbf{R}_J (\bar{H}^c + \mathbf{B}\alpha)$ are the joint locations for the given identity.
This formulation aligns with the standard structure of parametric models like MANO.

\section{Personalized Hand Template Generation}

\label{sec:PHT_generation}

The Personalized Volumetric Deformation Model (Sec.~\ref{sec:deform_model}) relies on a subject-specific Personalized Hand Template (PHT). This PHT provides the unique rest-state texture, geometry (surface and volume), skeleton, Linear Blend Skinning (LBS) weights, and a set of pose-dependent corrective blendshapes for an individual. A collection of PHTs are shown in Fig.~\ref{fig:hand_appearance}.

We generate this PHT for each subject by adapting a base generic parametric hand model, $\mathcal{H}$, parameterized by identity and pose. This base model comes with a pre-defined surface topology, kinematic skeleton, LBS weights, pose-dependent corrective blendshapes, and an associated base tetrahedral mesh (e.g., from TetGen~\cite{hang2015tetgen}) with fixed tetrahedral connectivity. Crucially, the PHT inherits these structural properties, i.e., surface topology, LBS weights, corrective blendshapes, and tetrahedral connectivity, from this generic model, ensuring consistency across subjects and with our deformation approach.

While our initial base model is a preliminary in-house version, it's important to note (as detailed in Sec.~\ref{sec:parametric_model}) that our full pipeline supports an iterative refinement loop: data registered using PHTs can be used to train an improved generic parametric hand model. This improved model can then serve as an even better base for re-generating more accurate PHTs. For the PHT generation described here, we assume access to such a suitable base generic model.

The geometry of the Personalized Hand Template (PHT) for each subject is derived by adapting our base generic parametric hand model to multiple input scans. Initially, we fit the generic model's identity and pose parameters to each scan, then refine this by solving for per-vertex surface offsets to capture fine scan details. This fitting is further improved through an iterative inverse solving process, which alternates between refining the parametric model's fit to the overall shape and re-calculating the offsets relative to this new parametric fit, ensuring accurate capture of the subject's intrinsic shape from each scan.

From each such personalized fit, we extract the subject's intrinsic shape relative to a canonical pose. To achieve a robust representation for the subject, these intrinsic shapes derived from multiple scans are aggregated using a RANSAC-based strategy to identify a consistent subset. The final subject-specific canonical identity and surface offset parameters are then determined by averaging across this reliable subset. This yields the PHT's personalized canonical surface and corresponding joint locations. Critically, to form the underlying personalized canonical tetrahedral mesh while preserving the consistent topology inherited from our generic template, we employ a tetrahedral As-Rigid-As-Possible (ARAP) deformation. This process fits the template's volume to the boundary defined by the personalized surface while maintaining internal rigidity.

The resulting PHT for the subject thus comprises this personalized canonical surface, canonical joint locations, inherited LBS weights, appropriately scaled pose corrective offsets, and the canonical tetrahedral mesh. For our experiments, this consistent PHT topology features 3,368 surface vertices, 6,698 surface faces, and an underlying volumetric mesh composed of 34,790 tetrahedral cells. This entire generation process is fully automated.

\paragraph{Texture Personalization}
We generate a single canonical texture map $T^{\text{canon}}$ for the PHT using the same RANSAC-selected subset of registered scans (from the geometry personalization step).
For each selected scan's registered mesh, we project input colors from all camera views onto its surface and aggregate them into a texture map $T_i$ in its specific UV space, using view-dependent weighting that favors normals oriented towards the camera.
Finally, these individual textures $\{T_i\}$ from the selected scans are robustly averaged to produce the final, high-quality canonical texture map $T^{\text{canon}}$ for the PHT.

\section{Details on Data Fidelity Terms} 
\label{sec:data_term_details}

Let $\mathcal{M}' = \{v'_i, n'_i\}$ denote the registered model's surface mesh with vertices $v'_i$ (Eq.~\ref{eq:lbs_final_posing}) and corresponding normals $n'_i$. Let $\mathcal{M}^t = \{v^t_j, n^t_j\}$ be the target mesh obtained from 3D reconstruction, with vertices $v^t_j$ and normals $n^t_j$.

\paragraph{3D Surface Alignment Term.}
When fitting to a target 3D mesh $\mathcal{M}^t$, we use a robust point-to-plane distance. For each vertex $v'_i$ on the registered mesh $\mathcal{M}'$, we find its $K$ nearest neighbors $\{v^t_j\}$ on the target mesh $\mathcal{M}^t$ within a search radius $\tau_d$. Let $\mathcal{N}_{\tau_d}^K(v'_i, \mathcal{M}^t)$ denote the set of indices $j$ for these neighbors.

We incorporate two masking strategies for robustness. First, a normal compatibility mask $m_{ij}^n$ ensures normals point in similar directions: $m_{ij}^n = \mathbf{1}[n'_i \cdot n^t_j > \tau_n]$, where $\tau_n$ is a threshold (e.g., $\tau_n=0$). This prevents matching front-facing to back-facing surfaces.

Second, a non-collision mask $m_i^c \in \{0, 1\}$ is applied to each registered vertex $v'_i$. This mask, defined in Sec.~\ref{sec:collision_reg} (Eq.~\ref{eq:col_pen_depth_based_topo}), reduces the influence of vertices involved in potential self-collisions, preventing distorted regions from dominating the fit. $m_i^c=0$ indicates a potential collision involving $v'_i$.

The robust 3D surface alignment term $E_{\text{3D}}$ is then:
\begin{equation}
E_{\text{3D}}(\mathcal{M}', \mathcal{M}^t) = \sum_i m_i^c \sum_{j \in \mathcal{N}_{\tau_d}^K(v'_i, \mathcal{M}^t)} m_{ij}^n \| n'_i \cdot (v'_i - v^t_j) \|_2^2.
\label{eq:3d_data_term_revised}
\end{equation}
This sums squared point-to-plane distances for valid, non-colliding correspondences.

\paragraph{Photometric Consistency Term.}
While 3D reconstructions provide valuable geometric information, they can sometimes be noisy, smoothed, or lack fine textural details present in the original images. To leverage this richer information from the raw 2D captures and provide additional constraints, we incorporate a 2D photometric consistency term. This term compares various attributes rendered from the registered mesh $\mathcal{M}'$ against corresponding target 2D attribute maps $I_{\mathrm{attr}}^{c}$ observed from or derived for camera view $c$. The primary attributes we consider are color, normals, depth and silhouette. We implement a differentiable rasterizer~\cite{cole2021differentiable}) that renders attribute $\mathrm{attr}$ of the registered mesh $\mathcal{M}'$ from viewpoint $c$, producing a rendered attribute image $I_{\mathrm{attr}}^{r,c}$

The target attribute maps $I_{\mathrm{attr}}^{c}$ are sourced as follows: for color attributes $I_{\mathrm{color}}^{c}$, we use the personalized canonical texture $T^{\text{canon}}$, detailed in Sec.~\ref{sec:PHT_generation} for registered mesh and directly use the raw captured RGB image from camera $c$; for geometric attributes, i.e., normal maps $I_{\mathrm{n}}^{c}$, depth maps $I_{\mathrm{d}}^{c}$, and silhouettes $I_{\mathrm{s}}^{c}$, we generate these target maps by rasterizing the 3D reconstruction (from Sec.~\ref{sec:reconstruction}) from camera view $c$.

To enhance robustness against discrepancies and focus the loss on reliable regions, we employ per-pixel masks. For instance:
\begin{enumerate}[label=\arabic*), itemsep=0pt, topsep=1pt, parsep=1pt] 
    \item A depth consistency mask $m^{d,c}_p = \mathbf{1}[|I_{\mathrm{d}}^{r,c}(p) - I_{\mathrm{d}}^{c}(p)| < \tau_d]$ can be used to down-weight pixels where the rendered depth significantly deviates from the target depth (from reconstruction), for a threshold $\tau_d$.
    \item A normal consistency mask $m^{n,c}_p = \mathbf{1}[I_{\mathrm{n}}^{r,c}(p) \cdot I_{\mathrm{n}}^{c}(p) > \tau_n]$ can be used to ensure alignment only where rendered and target normals are similar, for a threshold $\tau_n$.
    \item A general validity mask $m^{s,c}_p$ (typically derived from the target silhouette $I_{\mathrm{s}}^{c}$) ensures that the loss is computed primarily within the foreground region of interest.
\end{enumerate}
Let $M_p^c$ represent the combined per-pixel mask for pixel $p$ in camera view $c$, i.e., $M_p^c = m^{s,c}_p \cdot m^{d,c}_p  \cdot m^{n,c}_p  $ .
The robust 2D photometric consistency term $E_{\text{2D}}$ sums the squared differences over all relevant attributes, camera views $c$, and pixels $p$:
\begin{equation}
E_{\text{2D}}(\mathcal{M}', \{I_{\mathrm{attr}}^{c}\}_c) = \sum_{\mathrm{attr}} w_{\mathrm{attr}} \sum_{c,p} M_p^c(\mathrm{attr}) \cdot \| I_{\mathrm{attr}}^{r,c}(p) - I_{\mathrm{attr}}^{c}(p) \|_2^2.
\label{eq:2d_data_term_revised_v3}
\end{equation}
Here, $w_{\mathrm{attr}}$ is a weight for each attribute type, allowing us to balance their respective contributions. 

\paragraph{Landmark Alignment Term.}
Sparse landmark correspondences provide valuable constraints, especially for initialization and handling challenging poses. We define a landmark term encouraging alignment between landmarks $L'$ derived from the registered mesh $\mathcal{M}'$ and target landmarks $L^t \in \mathbb{R}^{K_L \times 3}$ ($K_L$ is the number of landmarks, e.g., 21). Target landmarks $L^t$ are obtained via lifting from 2D detections using a pre-trained network.
Landmark locations $L'$ on the registered mesh are computed as fixed linear combinations of its vertex positions $v'$, defined by a sparse matrix $R_{\text{Ldm}}$: $L' = R_{\text{Ldm}} v'$. The landmark term $E_{\text{Ldm}}$ minimizes the squared Euclidean distance:
\begin{equation}
E_{\text{Ldm}}(\mathcal{M}', L^t) = \| R_{\text{Ldm}} v' - L^t \|_2^2,
\label{eq:landmark_term_revised}
\end{equation}

\section{Physics-Inspired Regularization Terms} 
\label{sec:reg_term_details} 

Regularization is crucial to ensure plausible deformations, prevent artifacts like self-intersections or excessive distortion, and bridge gaps in noisy or incomplete data. As demonstrated in Fig.~\ref{fig:pyhsic_reg}, omitting key physical constraints can lead to undesirable outcomes such as surface collapse or inter-penetrations; our regularization strategy is designed to mitigate these issues. We employ surface, volumetric, collision, and temporal regularization terms.

\subsection{Surface Regularization}
\label{sec:surf_reg}

These terms operate on the posed surface mesh $\mathcal{M}' = \{v'_i\}$.

   \textbf{As-Rigid-As-Possible (ARAP) Energy} Preserves local shape details by penalizing deviations from rigidity~\cite{sorkine2007rigid}. It encourages the local transformation around each vertex $i$ in the deformed mesh $\mathcal{M}'$ to be close to a rotation $R_i$, relative to a reference mesh $\mathcal{M}_{\text{ref}} = \{v^{\text{ref}}_i\}$. 
    Let $\mathcal{M}_{\text{ref}} = v^c$. The energy is:
\begin{equation}
\begin{split}
E_{\text{ARAP-surf}}&(\mathcal{M}', \mathcal{M}_{\text{ref}}) = \\
&\sum_i w_i \sum_{j \in \mathcal{N}(i)} w_{ij} \| (v'_i - v'_j) - R_i (v^{\text{ref}}_i - v^{\text{ref}}_j) \|_2^2,
\end{split}
\end{equation}
    where $\mathcal{N}(i)$ is the 1-ring neighborhood of vertex $i$, $R_i$ is the optimal local rotation at vertex $i$, $w_{ij}$ are edge weights (typically cotangent weights~\cite{meyer2003discrete} derived from $\mathcal{M}_{\text{ref}}$), and $w_i$ is a per-vertex weight (e.g., local surface area) for discretization independence.

\paragraph{Laplacian Smoothing Energy} Encourages surface smoothness by penalizing the squared magnitude of the discrete Laplace-Beltrami vector at each vertex:
    \begin{equation}
    E_{\text{Lap-surf}}(\mathcal{M}') = \| L(\mathcal{M}') \|_F^2 = \sum_i \| L(v'_i) \|_2^2,
    \label{eq:lap_surf_revised} 
    \end{equation}
    where $L$ is the discrete Laplace-Beltrami operator. 

\subsection{Volume Regularization}
\label{sec:vol_reg}

To enforce physically plausible volumetric deformation and prevent element inversion or unnatural volume change, we apply regularization to the deformation induced by the tetrahedral offsets $\Delta^c$. These terms operate on the updated canonical tetrahedral vertices $\tilde{\rho}^c$, using the original canonical tet mesh $\rho^c$ as the reference state. Let deformed canonical tet mesh $\tilde{\mathcal{T}} = \{\tilde{\rho}_i^c\}$ and canonical tet mesh $\mathcal{T}^c = \{\rho_i^c\}$. 

\paragraph{Volumetric ARAP Energy} Analogous to surface ARAP, this encourages the local deformation within the volume (from $\mathcal{T}^c$ to $\tilde{\mathcal{T}}$) to be rigid:
    \begin{equation}
    E_{\text{ARAP-vol}}(\tilde{\mathcal{T}}, \mathcal{T}^c) = \sum_i w_i \sum_{j \in \mathcal{N}_{\text{tet}}(i)} w_{ij} \| (\tilde{\rho}_i^c - \tilde{\rho}_j^c) - R_i^{\text{vol}} (\rho_i^c - \rho_j^c) \|_2^2.
    \label{eq:arap_vol_revised}
    \end{equation}
    Here, $i, j$ index tetrahedral vertices, $\mathcal{N}_{\text{tet}}(i)$ are neighbors in the tet mesh $\mathcal{T}^c$, $R_i^{\text{vol}}$ is the optimal local rotation, and weights $w_i, w_{ij}$ are defined analogously to the surface term but on the tet mesh.

\paragraph{Neo-Hookean Elastic Energy} Models the tetrahedral mesh as an elastic material, strongly penalizing volume changes and inversions~\cite{bonet1997nonlinear}. The energy is summed over tetrahedra $k$:
    \begin{equation}
    E_{\text{NH-vol}}(\tilde{\mathcal{T}}, \mathcal{T}^c) = \sum_k V_k^c \cdot \phi_k(\tilde{\mathcal{T}}, \mathcal{T}^c),
    \label{eq:nh_vol_revised}
    \end{equation}
    where $V_k^c$ is the rest volume of tetrahedron $k$ in $\mathcal{T}^c$, and $\phi_k$ is the Neo-Hookean energy density for that tetrahedron:
    \begin{equation}
    \phi_k = \frac{\mu}{2} (\mathrm{Tr}(F_k^T F_k) - 3) - \mu \log J_k + \frac{\lambda}{2} (\log J_k)^2.
    \label{eq:nh_density_revised} 
    \end{equation}
    Here, $F_k \in \mathbb{R}^{3 \times 3}$ is the deformation gradient for tetrahedron $k$, mapping vectors from the rest state ($\mathcal{T}^c$) to the deformed state ($\tilde{\mathcal{T}}$). $J_k = \det(F_k)$ measures local volume change ($J_k=1$ preserves volume). $\mu$ and $\lambda$ are Lamé parameters controlling shear and volume resistance. This term effectively regularizes the tetrahedral offsets $\Delta^c$.

\subsection{Collision Prevention}
\label{sec:collision_reg}

Ensuring a physically plausible, collision-free hand registration is critical, especially given the hand's propensity for self-contact and complex articulation. We employ a collision prevention strategy that identifies inter-penetrating regions and applies a penalty term to encourage separation. This also informs a masking strategy for the 3D data term (Sec.~\ref{sec:data_term}).

\paragraph{Identifying Penetrating Vertices.}
We first identify vertices of the posed surface mesh $\mathcal{M}' = \{v'_i, n'_i\}$ that are likely inter-penetrating. A vertex $v'_i$ is considered "interior" or "penetrating" if a randomly cast ray originating from it intersects the mesh surface an odd number of times. Let $\mathcal{C}(\mathcal{M}')$ be the set of such identified penetrating vertices.

\paragraph{Finding Opposing Collision Candidates.}
For each penetrating vertex $v'_c \in \mathcal{C}(\mathcal{M}')$ with normal $n'_c$, we search for candidate collision points $v'_j$ on other parts of the surface. These candidates are typically found among the $K$ nearest neighbors to $v'_c$ on $\mathcal{M}'$ within a search radius $\tau_p$. We then apply two filters to refine these candidates:

\textit{Normal Filter ($m_{cj}^{n}$):} This filter identifies pairs where the surfaces are locally opposing, indicative of a collision. It is active if the dot product of their normals is negative:
    $m_{cj}^{n} = \mathbf{1}[n'_c \cdot n'_j < 0]$.
    This helps ensure that we are penalizing actual inter-penetrations rather than just proximity between surfaces facing the same way.

\textit{Topological Filter ($m_{cj}^{\text{topo}}$):} This filter excludes neighbors that are topologically close to $v'_c$ on the mesh surface. This is crucial to focus on collisions between distinct parts of the hand (e.g., different fingers, or a finger and the palm) rather than penalizing the natural proximity of adjacent vertices on the same surface. The filter is active if the geodesic distance on the mesh surface between vertex $c$ and vertex $j$, $\text{geodesic\_dist}(c, j)$, is greater than a threshold $R_{\text{topo}}$ (e.g., $R_{\text{topo}}$ could be equivalent to 3 edge hops):
    $m_{cj}^{\text{topo}} = \mathbf{1}[ \text{geodesic\_dist}(c, j) > R_{\text{topo}} ]$.

\paragraph{Collision Penalty Term ($E_{\text{Col-pen}}$).}
Based on the filtered candidate pairs (where $m_{cj}^n=1$ and $m_{cj}^{\text{topo}}=1$), we define a collision penalty term $E_{\text{Col-pen}}$. This term actively pushes penetrating surfaces apart:
\begin{equation}
E_{\text{Col-pen}}(\mathcal{M}') = \sum_{v'_c \in \mathcal{C}(\mathcal{M}')} \sum_{j} m_{cj}^{n} \cdot m_{cj}^{\text{topo}} \cdot \max(0, n'_c \cdot (v'_j - v'_c) + \epsilon_{pen})^2,
\label{eq:col_pen_depth_based_topo}
\end{equation}
where $j \in \mathcal{N}_{\tau_p}(v'_c, \mathcal{M}')$ are neighbors of $v'_c$ in the posed mesh within radius $\tau_p$, $\epsilon_{pen}$ is a small offset to ensure repulsion even at slight contact. The term $n'_c \cdot (v'_j - v'_c)$ approximates the penetration of $v'_c$ into the surface around $v'_j$ along $v'_c$'s normal. 

\paragraph{Non-Collision Mask for Data Term.}
Furthermore, we use these collision indicators to define the non-collision mask $m_i^c$ (used in Eq.~\ref{eq:3d_data_term_revised}) that down-weights the influence of potentially colliding vertices in the 3D data fidelity term. For any vertex $v'_i$:
\begin{equation}
m_i^c = 1 - \mathbf{1}[v'_i \in \mathcal{C}(\mathcal{M}') \land \exists j \text{ s.t. } m_{ij}^n \cdot m_{ij}^{\text{topo}} = 1].
\label{eq:collision_mask_def_topo_filter}
\end{equation}
This means if $v'_i$ is identified as penetrating AND it has at least one valid opposing collision partner $v'_j$ (that is not topologically adjacent), its influence in the $E_{\text{3D}}$ term is reduced.

\subsection{Skeletal Pose Regularization}
\label{sec:pose_reg}

To ensure anatomically plausible hand configurations, we directly regularize the hand's pose parameters $\beta$. These parameters $\beta = \{\beta_j\}_{j \in \text{Joints}}$ represent the Euler angle rotations for each joint $j$, defined in its local bone-oriented coordinate system.

The regularization term $E_{\text{Pose}}$ penalizes deviations from a canonical pose using a weighted L2 penalty:
\begin{equation}
E_{\text{Pose}}(\beta) = \sum_{j \in \text{Joints}} w_j \| \beta_j - \beta_j^{\text{pref}} \|_2^2.
\label{eq:pose_reg_l2_euler}
\end{equation}
The predefined per-joint weights $w_j$ allow for varying degrees of rotational freedom, discouraging unnatural joint contortions. This term acts as a prior, guiding the optimization towards more natural hand poses, particularly when input data is ambiguous or occluded.

\subsection{Temporal Regularization}
\label{sec:temporal_reg}

For registering sequences over time $t$, we enforce temporal consistency to ensure smooth motion and prevent jitter. Let $v'_i(t)$ be the position of the $i$-th posed surface vertex at time $t$, $\beta(t)$ the pose parameters, and $\Delta^c(t)$ the tetrahedral offsets. We apply finite difference (FD) penalties. Let $\Delta x(t) = x(t) - x(t-1)$ (1st order FD) and $\Delta^2 x(t) = x(t) - 2x(t-1) + x(t-2)$ (2nd order FD). 

The temporal regularization term $E_{\text{Temporal}}$ penalizes non-smooth changes in vertex positions, pose parameters, and volumetric offsets:
\begin{equation}
\begin{aligned}
E_{\text{Temporal}} &= \sum_t \Big( \lambda_{v} \sum_i \|\Delta^{(ord)} v'_i(t)\|_2^2 \\& + \lambda_{\beta} \|\Delta^{(ord)} \beta(t)\|_2^2 
 + \lambda_{\Delta} \|\Delta^{(ord)} \Delta^c(t)\|_F^2 \Big), 
\end{aligned}
\label{eq:temporal_reg_revised}
\end{equation}
where $ord \in \{1, 2\}$ selects first or second-order differences (penalizing velocity or acceleration), and $\lambda_{v}, \lambda_{\beta}, \lambda_{\Delta}$ are weights. This encourages smooth trajectories for the surface, skeleton, and intrinsic shape deformation.

\section{Parameters and Computational Performance} 
\label{sec:computation}

This section details key parameter values and computational performance for the neural reconstruction and volumetric registration stages of our VEPHand pipeline. A consistent set of parameters was utilized in our data processing, which proves to be robust and minimizes task-specific tuning. All distance-based parameters reported use meters (m) as the unit, and timings are based on an NVIDIA V100 GPU.
\subsection{Neural Reconstruction: Parameters and Performance}
\label{ssec:recon_details_performance}

For the Neural Reconstruction stage (Sec.~\ref{sec:reconstruction}), input images were processed at a resolution of 1536x1024, utilizing all 20 camera views. The ray integration bounds were set to $t_n=0.2$ and $t_f=5.0$. For scenarios employing our $L_{\text{contain}}$ regularization (typically narrow-FoV setups), its specific parameters were $w_{bg}=10^{-4}$, $s=10$, and $\rho_{allow}=0.1$. During mesh extraction, the Marching Cubes density iso-level was $\tau=50$.  Initial surface extraction used a coarse voxel size of $0.01$ within the foreground region, followed by a surface refinement step using a finer voxel size of $0.001$. For our proximity volume rendering during texturing, we sampled 128 vertices in a local neighborhood of $\epsilon=0.005$ along the surface normal.

Computationally, using multi-resolution hash grid encoding~\cite{muller2022instant}, the baseline reconstruction for a single frame typically converges in approximately 20-25 minutes (9000 iterations). Employing our optional segmentation-guided ray sampling strategy (Sec.~\ref{sec:acc_train}) further reduces this to approximately 10 minutes (3000 iterations) per frame, a speed-up of roughly 2x. The mesh extraction and texturing process adds approximately 30 seconds per frame.
\subsection{Volumetric Registration: Parameters and Performance}
\label{ssec:reg_details_performance}
For the Volumetric Registration stage (Sec.~\ref{sec:registration}), several key parameters were defined. In the 3D surface alignment term ($E_{\text{3D}}$), we used $K=5$ nearest neighbors, with a search radius $\tau_d$ that annealed from $0.05$ down to $0.01$ over optimization iterations, and a normal compatibility threshold $\tau_n=0$. For the 2D photometric consistency term ($E_{\text{2D}}$), attribute weights were $w_{\text{color}}=1.0$, $w_{\text{normal}}=5.0$, $w_{\text{depth}}=1.0$, and $w_{\text{silhouette}}=1.0$. The depth consistency threshold was $\tau_d=0.005$, and normal consistency was $\tau_n=0.2$. The weight for the landmark alignment term ($E_{\text{Ldm}}$) was annealed during optimization. The Neo-Hookean elastic energy ($E_{\text{NH-vol}}$) used Lamé parameters $\mu=50$ and $\lambda=250$. For the collision penalty term ($E_{\text{Col-pen}}$), the search radius was $\tau_p=0.01$, the topological filter distance $R_{\text{topo}}$ was 3 edge hops, and the penetration offset $\epsilon_{pen}=0.0005$.

The one-time, automated generation of a PHT for a new subject (Sec.~\ref{sec:PHT_generation}) takes approximately 25 minutes. Per-frame registration optimization typically requires 3 minutes for a single hand and 4 minutes for two-hand scenarios. The initialization phase, including a learned mesh prediction (inference time: ~5 seconds per frame), is minor compared to the iterative optimization cost.



\end{document}